\documentclass[journal]{IEEEtran}
\usepackage{textcase}
\usepackage[tablename=TABLE]{caption}
\DeclareCaptionTextFormat{up}{\MakeTextUppercase{#1}}
\usepackage{lscape}
\usepackage{array}
\usepackage{graphicx}
\usepackage{multicol}
\usepackage{multirow}
\usepackage{amssymb}
\usepackage{pifont}
\usepackage{tikz}
\usepackage{verbatim}
\usepackage{array}
\usepackage{longtable}
\usepackage{supertabular}
\usepackage{amsmath}
\usepackage{epstopdf}
\usepackage{verse}
\usepackage{cite}
\usepackage{color}
\usepackage{lettrine}
\usepackage{hyperref}
\usepackage{soul}
\usepackage{comment}
\usepackage{verbatim}
\usepackage{authblk}
\usepackage{subcaption}
\usepackage{tikz}
\def\checkmark{\tikz\fill[scale=0.4](0,.35) -- (.25,0) -- (1,.7) -- (.25,.15) -- cycle;}
\usepackage{amsmath}
\usepackage{pifont}

\newcommand{\xmark}{\text{\ding{55}}}

\newcolumntype{a}{>{\columncolor{LightCyan}}c}

\begin{document}

\title{The Duo of Artificial Intelligence and Big Data for Industry 4.0: Review of Applications, Techniques, Challenges, and Future Research Directions}
\author{
    \IEEEauthorblockN{Senthil Kumar Jagatheesaperumal\IEEEauthorrefmark{1}, Mohamed Rahouti\IEEEauthorrefmark{2}, Kashif Ahmad\IEEEauthorrefmark{3}, Ala Al-Fuqaha\IEEEauthorrefmark{3}, Mohsen Guizani\IEEEauthorrefmark{4}}
    \\
    \IEEEauthorblockA{\IEEEauthorrefmark{1} Department of Electronics and Communication Engineering, Mepco Schlenk Engineering College, Sivakasi, Tamil Nadu, India.
    \\\ senthilkumarj@mepcoeng.ac.in} \\
    \IEEEauthorblockA{\IEEEauthorrefmark{2}Department of Computer and Information Science, Fordham University, Bronx, NY, USA.
    \\\ mrahouti@fordham.edu} \\
     \IEEEauthorblockA{\IEEEauthorrefmark{3}Information and Computing Technologies (ICT) Division, College of Science and Engineering (CSE), Hamad Bin Khalifa University, Doha, Qatar.
   \\\{kahmad, aalfuqaha\}@hbku.edu.qa} \\
   
   \IEEEauthorblockA{\IEEEauthorrefmark{4}Computer Science and Engineering Department, Qatar University, Doha, Qtar.
    \\\ mguizani@qu.edu.qa} \\
}


\markboth{IEEE Internet of Things Journal,~Vol.~ , No.~ ,~2020}%
{Shell \MakeLowercase{\textit{\textit{et al.}}}: Bare Demo of IEEEtran.cls for IEEE Journals}
\maketitle
\begin{abstract}
The increasing need for economic, safe, and sustainable smart manufacturing combined with novel technological enablers, has paved the way for Artificial Intelligence (AI) and Big Data in support of smart manufacturing. This implies a substantial integration of AI, Industrial Internet of Things (IIoT), Robotics, Big data, Blockchain, 5G communications, in support of smart manufacturing and the dynamical processes in modern industries. In this paper, we provide a comprehensive overview of different aspects of AI and Big Data in Industry 4.0 with a particular focus on key applications, techniques, the concepts involved, key enabling technologies, challenges, and research perspective towards deployment of Industry 5.0. In detail, we highlight and analyze how the duo of AI and Big Data is helping in different applications of Industry 4.0. We also highlight key challenges in a successful deployment of AI and Big Data methods in smart industries with a particular emphasis on data-related issues, such as availability, bias, auditing, management, interpretability, communication, and different adversarial attacks and security issues. \textcolor{black}{In a nutshell, we have explored the significance of AI and Big data towards Industry 4.0 applications through panoramic reviews and discussions.} We believe, this work will provide a baseline for future research in the domain.

\end{abstract}
\IEEEpeerreviewmaketitle

\section{Introduction}
\label{introduction}
\IEEEPARstart{I}{ntelligent} manufacturing and smart machines in modern industries are being evolved with the support of Artificial Intelligence (AI), Machine Learning (ML), and Big Data approaches. Even though sustainable utilization of these modern trends is a long way off, the Industry 4.0 revolution with the support of the Industrial Internet of Things (IIoT) and other supporting technologies are succeeding in smart factories providing high productivity with intelligent manufacturing. Manufacturing in smart industries involves data-driven decision making for improving the continuous production process and also involves modern robots for the sustainable processing of complex tasks. Driven by the trends in Industry 4.0, smart factories need to be enabled with high-speed connectivity between machines, secure devices, and machines with smarter solutions, to predict machine and product failures and make recommendations to save cost and time. Imparting AI technology in Industry 4.0 enhances manufacturing sectors to form a cluster of modern industries with sustainable vibrant ecosystems. 
\begin{figure}[!t]
\centering
\centerline{\includegraphics[height=22 cm, width=10cm]{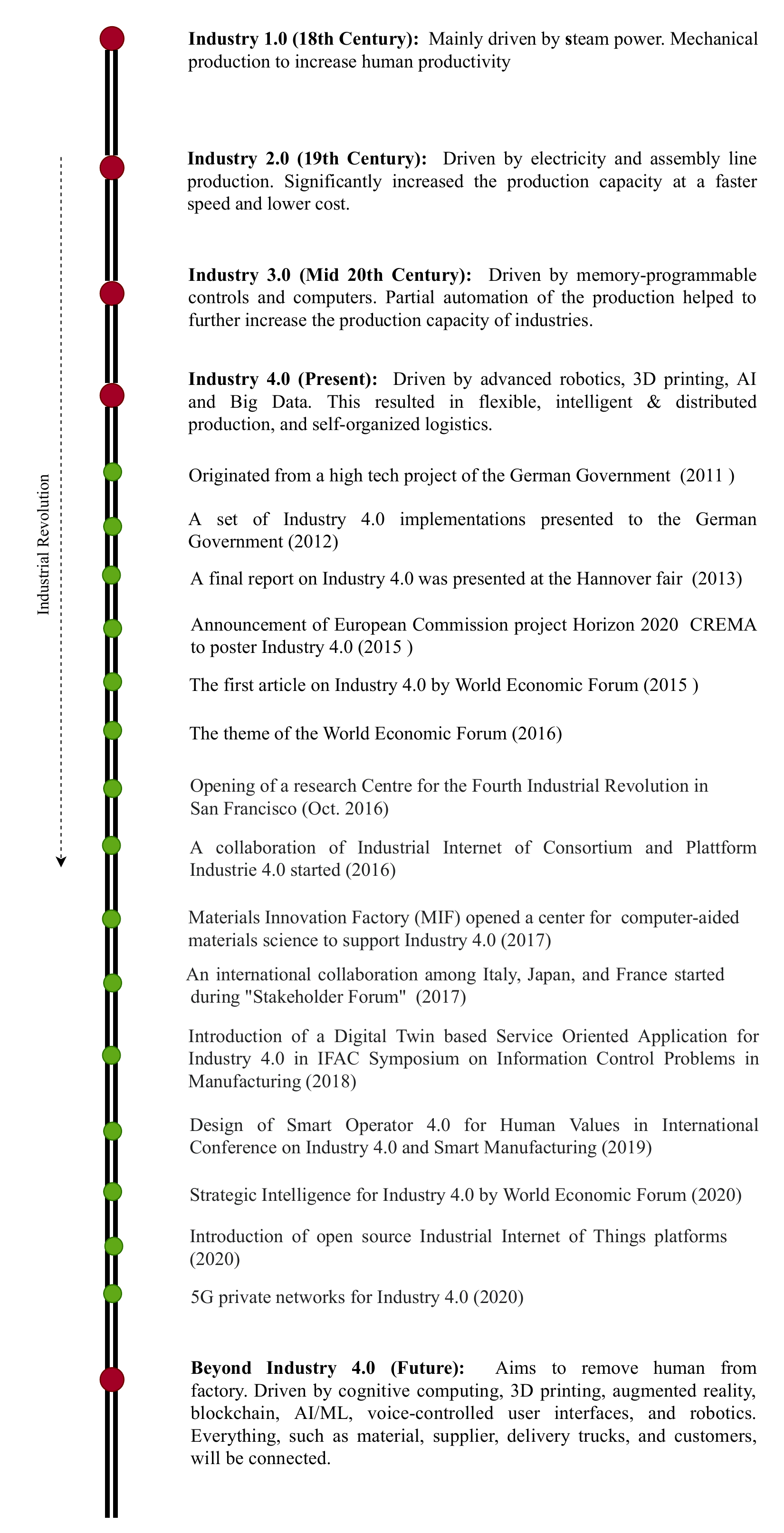}}
\caption{Timeline of industrial revolution. The red dots represent the revolution of Industry while the green dots represent different key developments in Industry 4.0.}
\label{fig:timeline}
\end{figure}

We are currently at the fourth industrial revolution (see Figure \ref{fig:timeline}), which is also known as the smart industry or Industry 4.0. This revolution started in Germany, which emphasizes the fact that every company needs certain healthy strategies to start with the smart industry. To this aim, the smart industry network consortium has developed a framework that gives companies guidance to determine where they stand, and where there are opportunities to implement smart industries. 



The fourth industrial revolution is about to boost industrial productivity by at least 30\% within a couple of years of full-fledged implementation. Industries have started using AI to reduce machine failure, improve quality control, increase productivity, and lower the costs of the products substantially\cite{angelopoulos_tackling_2020}. AI is equally useful for manufacturers and consumers. On one side, it makes the customers to be turned out as subscribers for new products and services. On the other side, it resulted in new opportunities for manufacturers, such as better customer service, easy maintenance, better tracking in logistics, and much more sophistication.  As a result, industries are enabled to have smart customer care services, smart dealerships, and smart experience centers for connecting customers. Moreover, immediate insights into the manufacturing sector and other services can be easily monitored and controlled by leveraging AI in industries. AI-based automation in industries also ensures quality services with the support of advanced robotics and 3D printing technologies. In short, there are several domains where AI plays a crucial role in Industry 4.0 as detailed in Section \ref{sec:AI_applications}. 

Moreover, in modern digital factories all systems, processes, machines, and products are digitally connected. This digitization of the industries results in a large collection of data covering different aspects of industries, which is a highly precious material for smart industries. The expansion and development of Big Data infrastructures in industries can generate a large amount of operational information that can no longer be interpreted by humans. There is already a focus on moving towards data analytic including the deployment of algorithms able to make autonomous and accurate decisions. In fact, the factory of the future is expected to consist of intelligent self-learning robots where the cyber-physical systems can communicate with each other. Machines themselves can request their demands, raw materials, and request services whenever needed. 

Despite the several advantages, digitization of the industries also brings several challenges, such as the availability, accessibility, and bias of data, data storage, management, data auditing, high-speed connectivity, security, and protection challenges. For instance, AI models in an industrial environment are at the risk of several security threats, where an adversarial attack, for example, launched by an attacker may disturb the predictive capabilities of the model. In addition, to develop stakeholders' trust in AI and Big Data techniques, the AI decisions should be interpretable.  

Moreover, we believe there is a deep connection between Big data and AI, and the duo can help industries to develop cognitive abilities. To fully exploit the potential of the data generated in smart industries, Big Data and AI should be deployed in an integrated manner. In this paper, we provide a detailed overview of opportunities and challenges in deploying AI and Big data in industries and also analyze the connection between the two technologies. 
\subsection{Scope of the Survey}
The main focus of the paper is on the role played by a duo of technologies namely AI and Big Data in industry 4.0. The paper pays particular attention to applications, techniques, and associated data-related, communication, and security challenges, and opportunities the duo brings in Industry 4.0. In detail, we provide a detailed overview of a wide range of AI and big data techniques and explore the Industry 4.0 applications that have benefited from AI and big data.  

The paper also identifies and discusses key technological, data-related, and security issues and challenges associated with a successful deployment of AI and Big Data in Industry 4.0. The paper also advises on the current limitations, pitfalls,  and future directions of research in Industry 4.0.  
\subsection{Related Surveys}

\begin{figure*}[!t]
\centering
\centerline{\includegraphics[height=16cm, width=19cm]{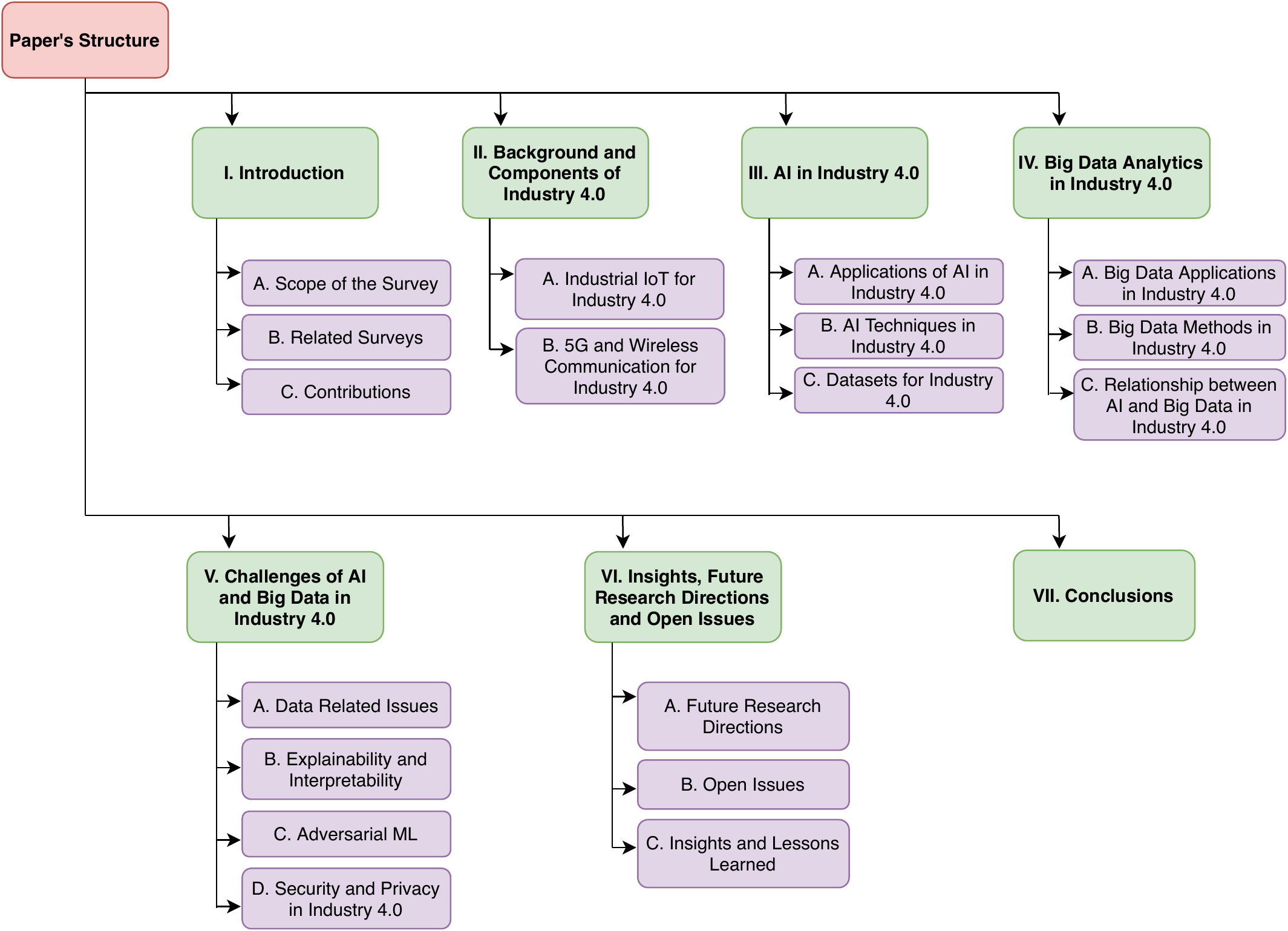}}
\caption{A Roadmap of the paper.}
\label{fig:paperStructure}
\end{figure*}

To the best of our knowledge, there is no prior work describing the junction of Industry 4.0 and the duo of AI and Big Data with particular attention to AI and Big Data techniques, applications, and associated data-related, technological and security challenges. However, few works have presented some common concepts, such as IIoT, Robotics, Big data, security, and other technologies used in Industry 4.0. For instance, Zhong et al\cite{zhong2017intelligent} reviewed several strategic plans for intelligent manufacturing using AI. Some of the AI-based recent developments in intelligent manufacturing are also addressed by Preuveneers \textit{et al.}\cite{preuveneers2017intelligent}. On the other hand, the authors in\cite{angelopoulos2020tackling}, reviewed several ML and IIoT-based solutions for smart industries. Table \ref{tab:survey} provides a summary of related works in terms of applications and challenges covered in each of them.  

However, common linkage of AI, smart manufacturing, IIoT, Robotics, Big data, and more importantly the data, security and communication challenges hindering a successful deployment of the duo of AI and Big Data technologies in Industry 4.0 are missing in the literature. This paper gives a particular attention to the linkage of these technologies and tries to fill up the potential research gap through the compilation of recent literature. 

\begin{table*}
\scriptsize
\caption{Summary of existing surveys related to artificial intelligence and Industry 4.0.} 
\label{tab:survey}
\begin{tabular}{|>{\arraybackslash}p{1.5cm}|>{\centering\arraybackslash}p{0.5cm}|>{\centering\arraybackslash}p{0.4cm}|>{\centering\arraybackslash}p{0.5cm}|>{\centering\arraybackslash}p{0.4cm}|>{\centering\arraybackslash}p{0.5cm}|>{\centering\arraybackslash}p{0.5cm}|>{\centering\arraybackslash}p{0.4cm}|>{\centering\arraybackslash}p{0.7cm}|>{\centering\arraybackslash}p{0.7 cm}|>{\arraybackslash}p{2.5cm}|>{\arraybackslash}p{4.5cm}|}
 \hline
\textbf{Refs. (Author)}
& \textbf{Year}
& \textbf{\tiny AI 4.0 \newline Review} 
& \textbf{\tiny Smart \newline Manufa-cturing}
& \textbf{\tiny IIoT \newline Review}
& \textbf{\tiny Big Data \newline Analytics}
& \textbf{\tiny Security \& Privacy}
& \textbf{\tiny Explain-ability}
& \textbf{\tiny Adversarial ML}
& \textbf{\tiny Datasets}
& \textbf{\tiny Big Data Issues}
& \textbf{\tiny Overview}
 \\ \hline \hline
Aceto \textit{et al.}\cite{aceto_survey_2019} & 2019 & \xmark{} & \xmark{} & \checkmark{} & \checkmark{} & \checkmark{} & \xmark{} &  \xmark{} & \checkmark{} & Distributed data management  & Analysis on the integration of key technology enablers for Industry 4.0 \\ \hline
Lu \textit{et al.}\cite{lu2017industry} & 2017 & \xmark{} & \xmark{} & \checkmark{} & \xmark{} & \checkmark{} & \xmark{} & \xmark{} & \xmark{} & Interoperability of industrial data & Examines the critical issues in interoperability and proposed a conceptual framework for Industry 4.0 \\ \hline
Zhong \textit{et al.}\cite{zhong2017intelligent}  & 2017 &\checkmark{} & \checkmark{} & \checkmark{} & \checkmark{} & \checkmark{} & \xmark{} & \xmark{} & \checkmark{} & Collection of real-time industrial data &  Strategic plans for intelligent manufacturing are highlighted and analyzed \\\hline
Oztemel \textit{et al.}\cite{oztemel2020literature} & 2020 & \xmark{} & \checkmark{} & \xmark{} & \checkmark{} & \xmark{} & \xmark{} & \xmark{} & \checkmark{} & Data assessment methodologies & With six design principles the conceptual framework of Industry 4.0 is analyzed \\ \hline
Xu \textit{et al.}\cite{xu2019big} & 2019 & \xmark{} & \xmark{} & \xmark{} & \checkmark{} & \checkmark{} & \xmark{} & \xmark{} & \checkmark{} & CPS data management & Surveys about research direction on complete autonomous operations in Industry 4.0 \\ \hline
Kamble \textit{et al.}\cite{kamble2018sustainable} & 2018 & \xmark{} & \checkmark{} & \checkmark{} & \xmark{} & \xmark{} & \xmark{} & \xmark{} & \checkmark{} & Guaranteed sustainability of data & Industry 4.0 framework is proposed for sustainable smart manufacturing \\ \hline
Pereira \textit{et al.}\cite{pereira2017review} & 2017 & \xmark{} & \checkmark{} & \checkmark{} & \xmark{} & \checkmark{} & \xmark{} & \xmark{} & \checkmark{} & CPS data management & Drives to develop new business models for Industry 4.0 considering various technological implications \\ \hline
Mittal \textit{et al.}\cite{mittal2018critical} & 2018 & \xmark{} & \checkmark{} & \xmark{} & \xmark{} & \checkmark{} & \xmark{} & \xmark{} & \xmark{} & Realistic maturity model & Impact and contribution of smart manufacturing in medium sized enterprises are summarized with smart manufacturing models \\ \hline
Alladi \textit{et al.}\cite{alladi2019blockchain} & 2019 & \xmark{} & \xmark{} & \checkmark{} & \xmark{} & \checkmark{} & \xmark{} & \xmark{} & \xmark{} & Security threats on industrial data & Challenges in blockchain implementation for Industry 4.0 in conjugation with IIoT are addressed with the application domains \\ \hline
Kerin \textit{et al.}\cite{kerin2019review} & 2019 & \xmark{} & \checkmark{} & \checkmark{} & \xmark{} & \checkmark{} & \xmark{} & \xmark{} & \checkmark{} & Value creating based on industrial data & Key observations from 29 research topics related to Industry 4.0 are summarized \\ \hline
Osterrieder \textit{et al.}\cite{osterrieder2020smart} & 2020 & \xmark{} & \checkmark{} & \checkmark{} & \xmark{} & \xmark{} & \checkmark{} & \xmark{} &\checkmark{} & Low generalizability of data & research model for smart manufacturing is put forward for future research in this domain \\ \hline
Davies \textit{et al.}\cite{davies2017review} & 2017 & \xmark{} & \checkmark{} & \xmark{} & \xmark{} & \xmark{} & \xmark{} & \xmark{} &  \checkmark{} & Lean environmental operations & Importance of socio-technical requirements for lean manufacturing are reviewed \\ \hline
Khan \textit{et al.}\cite{khan2016survey} & 2016 & \xmark{} & \checkmark{} & \xmark{} & \xmark{} & \xmark{} & \xmark{} & \xmark{} & \checkmark{} & Production level management & New technology and tools are introduced to support the growth of Industry 4.0 \\ \hline
Trappey \textit{et al.}\cite{trappey2017review} & 2017 & \xmark{} & \checkmark{} & \checkmark{} & \xmark{} & \xmark{} & \xmark{} & \xmark{} & \checkmark{} & Data based product life cycle management & Impact of IoT standards for enhancing the productivity in smart manufacturing are reviewed with case studies \\ \hline
Yli \textit{et al.}\cite{yli2019adapting} & 2019 & \xmark{} & \checkmark{} & \checkmark{} & \xmark{} & \checkmark{} & \xmark{} & \xmark{} & \checkmark{} & Matured data sharing architectures & Communication and networking standards for agile manufacturing are summarized \\ \hline
Sanchez \textit{et al.}\cite{sanchez2020industry} & 2020 & \xmark{} & \checkmark{} & \checkmark{} & \xmark{} & \xmark{} & \xmark{} & \xmark{} & \checkmark{} & Data integration issues & Collaboration, organization and system integration features for Industry 4.0 are discussed \\ \hline
Preuveneers \textit{et al.}\cite{preuveneers2017intelligent} & 2017 & \checkmark{} & \checkmark{} & \xmark{} & \checkmark{} & \checkmark{} & \xmark{} & \xmark{} & \checkmark{} & Data protection safeguards & Recent developments and challenges in intelligent manufacturing are addressed \\ \hline
Kim \textit{et al.}\cite{kim2017review} & 2017 & \xmark{} & \xmark{} & \xmark{} & \checkmark{} & \checkmark{} & \xmark{} & \xmark{} & \checkmark{} & CPS data management & Imparting data security in CPS systems for Industries are reviewed \\ \hline
Sony \textit{et al.}\cite{sony2020critical} & 2020 & \xmark{} & \checkmark{} & \xmark{} & \xmark{} & \xmark{} & \xmark{} & \xmark{} & \checkmark{} & Guranteed sustainability of data & Critical factors are addressed for sustainable implementation of Industry 4.0 \\ \hline
Li \textit{et al.}\cite{li2017review} & 2017 & \xmark{} & \checkmark{} & \xmark{} & \xmark{} & \xmark{} & \xmark{} & \xmark{} & \xmark{} & Qualtiy of data among industrial networks & Design challenges in wireless networking for smart industries are discussed \\ \hline
Souza \textit{et al.}\cite{souza2020survey} & 2020 & \xmark{} & \xmark{} & \checkmark{} & \xmark{} & \checkmark{} & \xmark{} & \xmark{} & \checkmark{} & Data reliability issues & From the context of Industry 4.0, how strategic decisions affect the reliability of products are summarized \\ \hline
Leng \textit{et al.}\cite{leng2020blockchain} & 2020 & \xmark{} & \checkmark{} & \xmark{} & \xmark{} & \checkmark{} & \xmark{} & \xmark{} & \xmark{} & Guaranteed sustainability of data & Successful empowerment of blockchain technology amidst the challenges in industries are addressed \\ \hline
Nunez \textit{et al.}\cite{nunez2020information} & 2020 & \xmark{} & \checkmark{} & \xmark{} & \xmark{} & \xmark{} & \xmark{} & \xmark{} & \checkmark{}& Lean supply chain data management & Impact of digital technology supporting the policymakers for successful implementation of modern trends in industries are reviewed \\ \hline
Calabres \textit{et al.}\cite{calabrese2020evolutions} & 2020 & \xmark{} & \checkmark{} & \xmark{} & \xmark{} & \xmark{} & \xmark{} & \xmark{} & \checkmark{} & Industrial process monitoring & Evolution and revolution aspects of Industry 4.0 are addressed for foreseeing the business prospects \\ \hline
Angelopoulos \textit{et al.}\cite{angelopoulos2020tackling} & 2020 & \checkmark{} & \xmark{} & \checkmark{} & \xmark{} & \checkmark{} & \xmark{} & \checkmark{} & \checkmark{} & CPS data management & ML based closed loop operation of human and machine are analyzed for using IIoT \\ \hline
Nguyen \textit{et al.}\cite{nguyen2020systematic} & 2020 & \xmark{} & \xmark{} & \xmark{} & \checkmark{} & \checkmark{} & \xmark{} & \xmark{} & \checkmark{} & Data privacy implications & Handling of Big data effectively in oil \& gas industries are addressed with appropriate security measures \\ \hline
ur Rehman \textit{et al.}\cite{ur_rehman_role_2019} & 2019 & \xmark{} & \checkmark{} & \checkmark{} & \checkmark{} & \xmark{} & \xmark{} & \xmark{} & \checkmark{} & Imparting intelligence on IIoT data & Case studies on IIoT and usage of big data analytics frameworks were explored. \\ \hline
Kumar \textit{et al.}\cite{kumar2020adversarial} & 2020 & \xmark{} & \xmark{} & \xmark{} & \xmark{} & \checkmark{} & \xmark{} & \checkmark{} & \checkmark{} & Data poisoning issues & Security enforcement using adversarial ML in smart industries are enumerated.  \\ \hline
Guidotti \textit{et al.}\cite{guidotti2018survey} & 2018 & \xmark{} & \xmark{} & \xmark{} & \checkmark{} & \xmark{} & \checkmark{} & \xmark{} & \checkmark{} & Trust issues in data & Provides explainable methods that could be used across multiple industries \\ \hline
Our Survey & 2021 & \checkmark{} & \checkmark{} & \checkmark{} & \checkmark{} & \checkmark{} & \checkmark{} & \checkmark{} & \checkmark{} & Industrial data related issues are addressed & Focuses the integration of AI with Smart manufacturing, IIoT, Big data, Security and regulatory standards \\ \hline
\end{tabular}
\end{table*}
\subsection{Contributions}

This paper focuses on a duo of modern technologies (AI and Big Data) playing an important role in industry 4.0. The paper aims to explore the potential applications, opportunities, and challenges associated with the deployment of the two technologies in smart industries.  

The contributions of this survey can be summarized as follows:

\begin{itemize}
\item We provide a detailed overview of the existing literature on the use of AI and Big data in industry 4.0 with a particular focus on enabling technologies, techniques, applications, and associated challenges and opportunities. 

\item We pay particular attention to different types of challenges associated with AI and Big data in smart industries including data-related challenges (e.g., the availability, accessibility, storage and management, auditing, and transmission issues), security \& privacy, interpretation (explainability), and adversarial attacks on AI models. 

\item Being a key component of industry 4.0, we also provide an overview of IIoTs and analyze how insights are extracted from data generated in IIoTs through AI and Big data techniques.

\item We also discuss and try to explore the connection between the duo of technologies, and how they are reshaping the industries when integrated.

\item We then identify the limitations,  pitfalls,  and open research challenges in the domain.

\item We also provide an overview of different types of datasets that can support future research in the domain. 
\end{itemize}

The roadmap of the paper is depicted in Figure \ref{fig:paperStructure}. Precisely, Section \ref{sec:background} discusses the key components and enabling technologies of industry 4.0 and provides a detailed description of how IIoTs are helping in industries. Section \ref{AIindustry} provides an overview of the literature on AI in industry 4.0 with a particular focus on applications and techniques. Section \ref{bigdata} focuses on Big Data for industry 4.0. Section \ref{sec:challenges} highlights the key challenges in the deployment of AI and Big Data techniques in industry 4.0. In Section \ref{insights}, we provide key insights and lessons learned from the literature and future research directions. Finally, Section \ref{conclusion} concludes the paper. All the related acronyms used in this article are listed in Table \ref{tab:terms}.

\begin{table}[t]
\centering
\caption{A Summary of acronyms used in this article.}\vspace{9pt}
\label{tab:terms}
\begin{tabular}{|p{2cm}|p{6cm}|}
\hline
Acronym     & Description 								\\ \hline \hline
AI          & Artificial Intelligence           		\\ \hline
AMCL        & Adaptive Monte-Carlo Localization         \\ \hline
ANN         & Artificial Neural Networks          		\\ \hline
AR          & Augmented Reality        		            \\ \hline
CBRS        & Citizens Broadband Radio Service          \\ \hline
C-RAN       & Centralized Radio Access Network          \\ \hline
CV          & Computer Vision		    				\\ \hline
DoS         & Denial of Service                         \\ \hline
DDoS        & Distributed Denial of Service             \\ \hline
DIKW        & Data–Information–Knowledge–Wisdom         \\ \hline
FiWi        & Fiber-Wireless                            \\ \hline
GAN         & Generative Adversarial Network            \\ \hline
GB          & Gradient Boosting          		        \\ \hline
HMI         & Human Machine Interaction                 \\ \hline
HTC         & Human Type Communication                  \\ \hline
IAI         & Industrial Artificial Intelligence        \\ \hline
ICA         & Independent Component Analysis         	\\ \hline
ICT         & Information Communication Technology	    \\ \hline
IIoT        & Industrial Internet of Things 		 	\\ \hline
IMT         & International Mobile Telecommunications   \\ \hline
IT          & Information Technology                    \\ \hline
KNN         & K-Nearest Neighbor          		        \\ \hline
LTE         & Long-Term Evolution                       \\ \hline
ML          & Machine Learning		                    \\ \hline
DL          & Deep Learning		                        \\ \hline
MLP         & Multi-layer Perceptions                   \\ \hline
MPC         & Model Predictive Controller		        \\ \hline
NFV         & Network Function Virtualization           \\ \hline
NLP         & Natural Language Processing               \\ \hline
PdM         & Predictive Maintenance                    \\ \hline
PMML        & Predictive Model Markup Language          \\ \hline
PCB         & Printed Circuit Board                     \\ \hline
PSO         & Particle Swarm Optimization               \\ \hline
QoS         & Quality of Service                        \\ \hline
RF          & Random Forest		                        \\ \hline
RFID        & Radio Frequency Identification		    \\ \hline
SCADA       & Supervisory Control and Data Acquisition  \\ \hline
SDN         & Software-Defined Networking               \\ \hline
SP          & Service Providers        		            \\ \hline
SVM         & Support Vector Machine          		    \\ \hline
WSN         & Wireless Sensor Network                   \\ \hline
XAI         & Explainable Artificial Intelligence       \\ \hline
\end{tabular}
\end{table}

\section{Background and Components of Industry 4.0}
\label{sec:background}
\begin{figure*}
  \centering    
  \includegraphics[width=\textwidth]{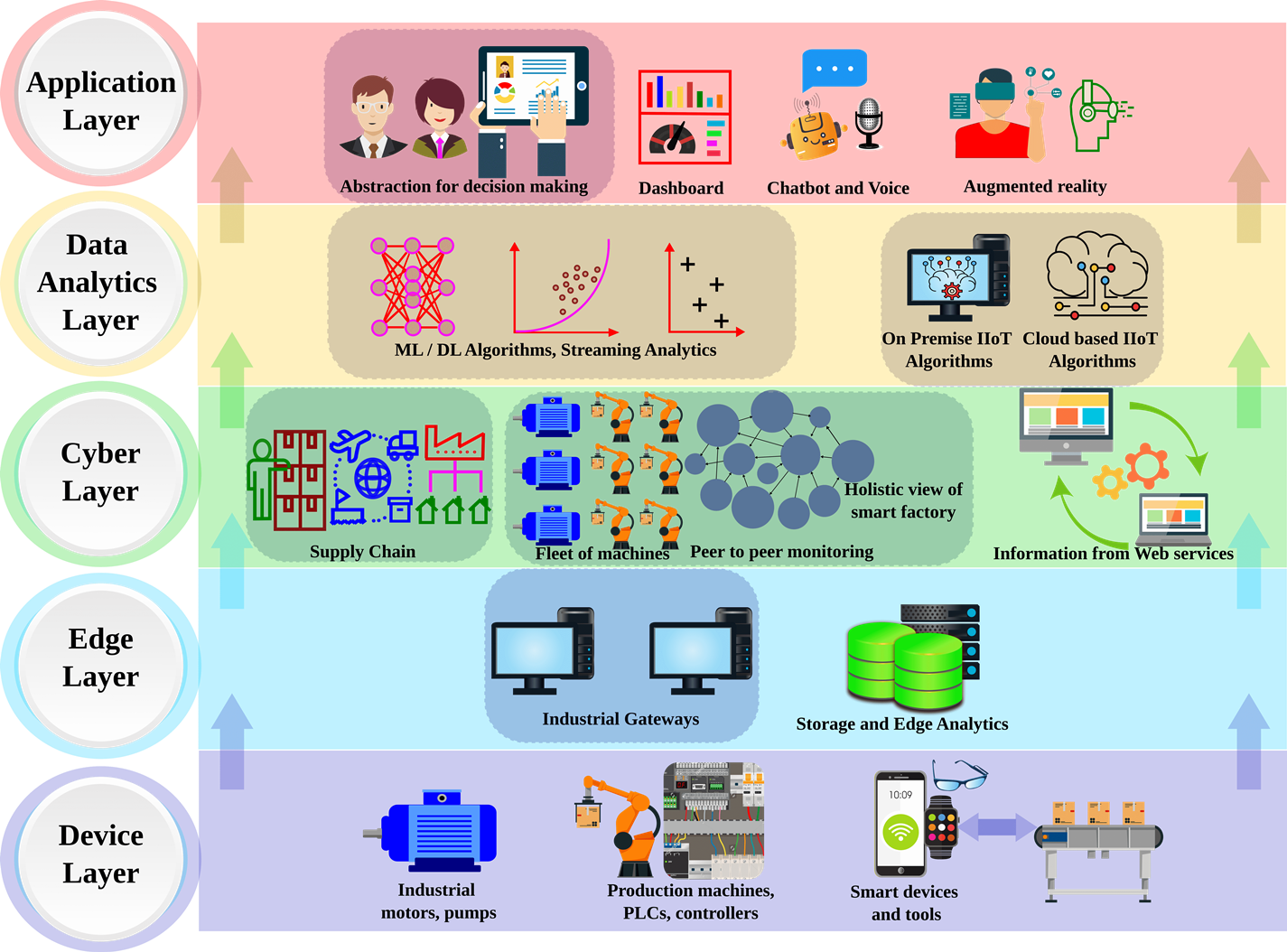}
  \caption{Layers of operations in smart industries with Industry 4.0 framework.}
     \label{fig:Ind40layers}
\end{figure*}

In this section, we begin with a brief introduction to Industry 4.0, highlighting the basic components and principles along with the key advantages that lead to their success. 
As shown in Figure \ref{fig:Ind40layers}, smart machines and devices in the industry 4.0 ecosystem fall under five layers of operations, namely (i) device layer, (ii) edge layer, (iii) cyber layer, (iv) data analytics layer, and (v) application layer. The device layer includes the machines, robots, PLC, controllers, and smart wearable devices employed for controlling the industrial devices. The edge layer stores the data acquired from the industrial machines through the physical layer communication protocols, such as ZigBee, wifi, and Bluetooth. The cyber layer manages the supply chain and handles information from web services. It also engages the networking of machines with trusted services and solutions. The data analytics layer runs certain ML algorithms and performs cloud analytics on the IIoT data.  

In order to fully exploit the benefits of the smart industries by implementing the industry 4.0 framework, these layers especially the application layer can be enabled with several components, such as a dashboard, user visualization resources, augmented reality, and chat-bots. 
The following are the vital components of the Industry 4.0 framework:

\begin{itemize}

    \item \textit{Sensors And Connected Devices}: Industrial components with sensors, hardware, software, and other integrated services are designed to enable the users to enjoy convenient, sustainable, and secure means of operations. 
    
    \item \textit{Big Data}: Smart machines connected in industries are capable of generating a huge amount of data. In smart industries, data is collected from several sources, such as smart machines, different plants, and the products manufactured in the plants. Industrial Big Data could be used separately and can be linked to other platforms, such as mobile/edge devices, cloud services, and on-premise systems. 

    \item \textit{The Cloud}:  Cloud is an integral component of Industry 4.0, and connected industries through cloud services are becoming a key part of the connected world with the support of Industry 4.0. Every salient feature of smart industries can be made available to a larger audience by leveraging the orchestration capabilities of cloud services.
        
    \item \textit{Augmented Reality}: With digitization being the vital driving force for technological trends and new developments, networking of machines and gathering information from them including basic and status data are of high importance.  Augmented reality in smart industries utilizes this data and databases with animations and 3D models on every tool the workspace offers virtually, allowing the toolmakers to facilitate processes and reduce the error rate significantly. 
    
    \item \textit{AI}:  Industrial applications by harnessing the power of AI reduce the machine failure rate, improve quality control, reduce costs involved, and also improves productivity. AI also creates new opportunities for the manufactures in terms of improved customer services, maintenance, marketing, and logistics.
    
    \item \textit{Digital Twin}: digital twin represents a virtual model monitoring an object or process throughout its life-cycle in smart industries.  For enhancing the business outcomes of smart industries, digital twins could play a crucial role. They are living models, which acquire environmental as well as operational data and constantly update themselves by learning from the data. It can also predict failures, unplanned outages, forecast opportunities, and also helps to optimize or provide mitigation events. 
      
    \item \textit{Cybersecurity}:  Privacy, trustworthiness, and other allied security issues of the industrial data are interlinked entities when considering cyber-physical industrial systems. They act as a protective shield for the smart machines connected to the internet from cyberattacks. Overall harm to the machines, hardware, and software components can be eliminated by deploying robust cybersecurity systems in smart industries.
        
    \item \textit{Additive Manufacturing and Digital Scanning}: As the end customer preferences are changing day by day, instead of employing skilled laborers for each taste of end-users, additive manufacturing provides a lightweight solution. It provides durable and proven solutions based on the demands of the customers by replacing metals with 3D products developed using stronger plastic components. 3D printing solutions provide developers to iterate and come up with ideas for much faster-proven solutions in additive manufacturing.
\end{itemize}


\subsection{Industrial IoT for Industry 4.0}
\label{IIoT}

The automation of industrial systems can be accomplished through smart devices that are capable of sensing the industrial environment, collecting valuable information, processing the data, and establishing effective communication served through IIoT. In\cite{khan2020industrial}, IIoT is envisioned as the new domain of IoT specifically used for industrial automation and also highlighted as one of the enabling technologies of Industry 4.0. The focus of IIoT is connecting devices/machines in smart industries to the internet, to each other, and to the workers. A typical IIoT solution for smart industries includes devices/machines, internet connectivity, communication setup, data, cloud and edge deployments, data analytics, and intelligence processing of the data to take appropriate actions. There are several applications/use-cases of IIoTs in Industry 4.0. 

\begin{figure}[!ht]
  \centering    
  \includegraphics[width=0.40\textwidth]{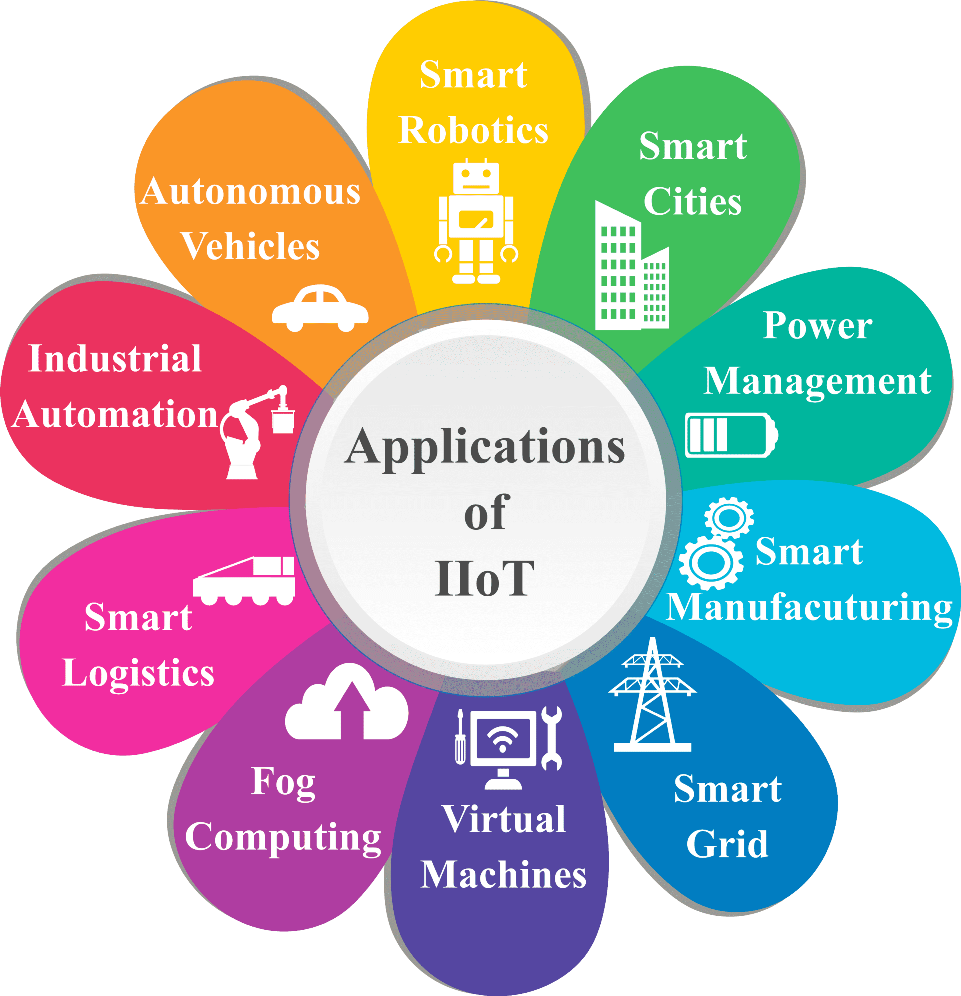}
  \caption{Major Applications of IIoT in Industry 4.0 framework.}
     \label{fig:IIoTAppl}
\end{figure}

For instance, real-time status monitoring in production lines of manufacturing processes is one of the primary requirements for making appropriate decisions on the quantity and quality of products to be marketed. Such analyses and decisions are made based on heterogeneous IIoT data of larger volume from production lines with the support of other technologies, such as wireless sensor networks and radio frequency identification (RFID) technology\cite{chen_intelligent_2020}. Such applications/use-cases of IIoTs demonstrate the effectiveness of IIoT-based solutions for monitoring and analysis of different tasks in the production line. Similarly, one of the possible mechanisms of better task scheduling for delay insensitive applications can be achieved using collaborative scheduling between multiple clusters of tasks\cite{xia_resource_2020}. This mechanism can be derived for IIoT applications with better task scheduling strategy implementation. There are other several interesting applications of IIoTs in Industry 4.0. Figure \ref{fig:IIoTAppl} summarizes some key applications of IIoTs in Industry 4.0.  

The integration of Big Data with the IIoT technology is indispensable for processing a large volume of data. However, they are also prone to challenges due to the storage, computation, and processing of large amounts of data \cite{ur_rehman_role_2019}. A detailed description of different challenges associated with the data collected through IIoT is provided in Section \ref{sec:challenges}. 

\subsection{5G and Wireless Communication for Industry 4.0}

\begin{figure}[!ht]
  \centering    
  \includegraphics[width=0.45\textwidth]{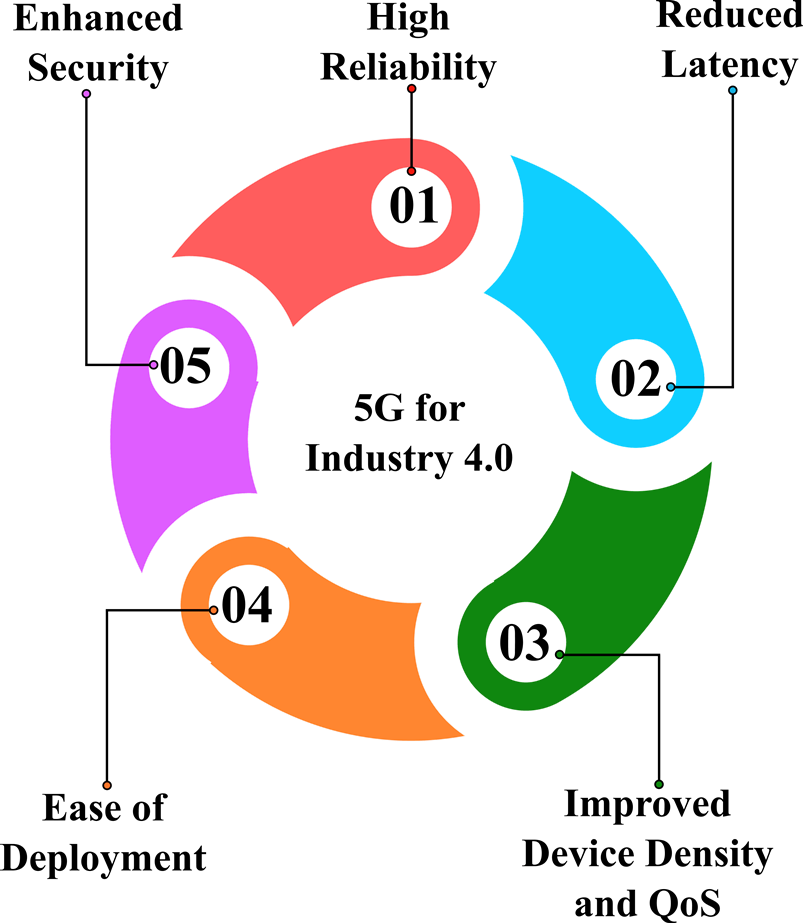}
  \caption{Impact of 5G services in Industry 4.0 applications.}
     \label{fig:5G40}
\end{figure}

5G communication standard not only enables humans to communicate and download data from the internet faster but also allows devices, such as sensors and other machines in an IoT environment \cite{5G}. The implementation of 5G services for Industry 4.0 results in highly reliable smart machines, with reduced latency, improves QoS, easy deployment, and enhanced security as depicted in Figure \ref{fig:5G40}.

In Smart industries, manufacturing has evolved from simple connections for communication between industrial machines. They require high-speed communication to collect and analyze data for driving dynamic intelligent actions. Quicker analysis of industrial data and accurate decision making based on the data are made feasible when the 5G communication systems are being paired with AI and Big Data. Smart factories use Big Data from interconnected devices driven by AI and 5G communications for establishing better ecosystems for connected machines in the industries.\cite{garg_guest_2020}. The striking features of 5G overcome the challenges in current communication technologies and promise to be a key enabler for smart factories\cite{rao_impact_2018}. 
According to the 5G Alliance for connected industries and automation, 5G communication technology will uplift the growth of industry 4.0 to the next level. Moreover, though the mobile technologies and industrial revolutions have happened separately, the 5G factory of the future brings them together to fulfill industrial requirements that cannot be fulfilled by either of them alone. This vision can be realized by slicing the physical 5G network into isolated logical networks. In this regard, Barakabitze \textit{\textit{\textit{et al.}}}\cite{barakabitze_5g_2020} reviewed and provided solutions for softwarization and slicing using Network Function Virtualization (NFV) and Software-Defined Networking (SDN) respectively in 5G networks. 

The Fiber-Wireless (Fi-Wi) network access in 5G communication overcomes the challenges in human type communication (HTC) by providing high bandwidth through the use of optical networks\cite{liu_virtual_2019}. Moreover, the traffic reallocation problems in the networks are also overcome using virtual network embedding algorithms and Q-Learning for managing the IIoT traffic.

\section{AI for Industry 4.0}
\label{AIindustry}
Nowadays, many industries have started harnessing the power of AI to a larger extent. In the focus of addressing the challenges and transforming towards Industry 4.0, AI technology is used for reducing the chances of machine failure, improving the product quality control, increasing productivity in industries, substantial cost reduction of products and thereby increasing the potential users of the product in the market. AI provisioning is an important feature, thanks also to the degree of flexibility provided by the learning frameworks, as it will help the AI paradigm to set itself with a primary role in the deployment of future smart industries that deploy Industry 4.0 standards. 

This section answers the question, where should AI frameworks be deployed in smart industries. To answer this question, we cover the following key aspects: (i) AI-based Industry 4.0 use cases (ii) deployment of AI techniques in Industry 4.0 applications, and (iii) datasets available for training and evaluation of AI models in smart industries. 

In the next subsection, we discuss some key applications of AI in Industry 4.0.

\subsection{Applications of AI in Industry 4.0}
\label{sec:AI_applications}

Some key applications of AI in industry 4.0 are provided below.

\subsubsection{Predictive Quality and Yield} In smart industries, AI frameworks aid in predicting the quality of the products throughout their life-cycles, which ultimately improve the yields to a larger extent. The quality of the products, which is what the customer demands, can be analyzed using predictive analytics via improved forecasting methods on existing products. Quality 4.0, which is a quality management subsystem in industry, for ensuring quality products in Industry 4.0\cite{lee2019quality}. In smart industries, with an abundance of quality data collected from different smart industrial machines, advanced analytics becomes highly feasible to effectively predict product quality and yield. Quality prediction of products at different stages of the life-cycle and real-time monitoring helps in ensuring improved production yield, better quality management, and improved customer value. 

The quality control loop in smart industries involves the iterative process of predicting products' yield. The predictive quality analysis helps in establishing baseline performance measures, monitoring the actual performance of smart machines, comparing and analyzing the performance with benchmark standards, and taking different actions accordingly. Lee \textit{et al.}\cite{lee2019quality}, presented an extensive survey on predictive quality management by leveraging AI, Big Data, and smart sensors in Industry 4.0 applications. Implementation of predictive quality management requires cooperative efforts from all the stakeholders of smart industries. They are highly helpful in estimating and figuring out what kind of quality issues are likely to occur when the product is out in the market.  
   
\subsubsection{Predictive Maintenance}
Predictive maintenance in smart industries is highly beneficial for industrial business executives, equipment maintenance managers, and the workers employed in modern industries. It is used to predict the probability of occurrence of a failure in smart machines and maintenance can be carried out in advance. Industry 4.0 compliance small-medium enterprises involved in a study of analyzing the value and cost factors encountered in the CNC machines and their predictive positive impacts show better economic and life cycle improvements of the products\cite{adu-amankwa_predictive_2019}.

The literature shows that managing smart industrial machines in run-time with the support of cyber-physical systems and IIoT plays a crucial role in Industry 4.0. Fitting the above-mentioned paradigm, a distributed algorithm is developed in\cite{upasani_distributed_2017} to plan the maintenance of machines in a decentralized manner with run-time analysis. The presented framework\cite{upasani_distributed_2017} provides intelligence-based maintenance scheduling for identically parallel multi-component machines over a job shop-enabled scenario of manufacturing. The intelligent algorithm in this framework is integrated within a CPS-IIoT paradigm and has a runtime scale following  the number of deployed machines. The performance efficiency of this solution was demonstrated over conventional centralized heuristics- namely, Memetic Algorithm and Particle Swarm Optimization.

In another work by O'Donovan \textit{et al.}\cite{odonovan_comparison_2019}, real-time performance monitoring of the machines and their reliability are compared between conventional cloud computing and fog computing approaches. It is targeted for Industry 4.0 applications with embedded ML and its latency, failure rates are analyzed under different scenarios. More remarkably, this research\cite{odonovan_comparison_2019} concluded that while reliability and latency in CPS interfaces designed with conventional/centralized cloud computing and decentralized computing (i.e., fog) technologies to enabled real-time intelligent learning solutions for I 4.0, communication failures rate can reach 6.6\% under different levels of communication stress in the cloud interfaces.

Moreover, delay encountered in predictive maintenance due to migration in an IIoT scenario is effectively addressed in\cite{wang_migration_2020}, using a cooperative strategy of information exchange in the developed cloud computing migration mode. Precisely, to improve resource utilization, the proposed approach leverages a partitioning mechanism based on a cyclic fine-grained directed graph and uses the concept of supply-demand similarity to compute the similarity level between the amount of computing node resources and requirements of task computing (at the edge computing platform). In\cite{liu_industrial_2020}, proactive maintenance and accurate tracking of products during the production stage are implemented using secure industrial blockchain-based services for effective product life cycle management. Similarly, in\cite{hehenberger_design_2016}, condition monitoring and predictive maintenance are implemented for the wind turbine system and the analysis of subsystems are performed using cloud services. The migration process towards future industries needs to adapt with service-oriented operations utilizing IIoT, cloud computing, and different levels of automated control of machines in smart industries\cite{breivold_towards_2020}.
\subsubsection{Smart Manufacturing}
The integration of operational and information technologies in modern industries is achieved through IIoT, AI, data analytics, Industry 4.0, and smart manufacturing processes. Smart manufacturing tasks allow machines to monitor physical processes during manufacturing and use the data acquired from IIoT devices in the factories to make predictive, corrective, and adaptive decisions with improved operational costs. Industry 4.0 defines smart manufacturing as fully integrated, collaborative manufacturing systems that respond in real-time to meet changing demands and conditions in the factory, supply chain, and customer requirements.  


In the literature, several approaches have been proposed for smart manufacturing. For instance, agent-based architectures for smart manufacturing include smart agents and cloud-assisted services providing collaborative negotiation and appropriate decision making\cite{tang_casoa_2018}. The agents involved in the architecture reduce the complexity of the framework, thereby increasing the robustness and adaptiveness while dealing with multiple product manufacturing processes. Similarly,  Digital twin-driven\cite{leng_digital_2020,lu_digital_2020} approach automates the manufacturing process by defining fixed standards, tools with open architecture providing flexible reconfiguration capability. The Digital Twin technology drives a smart manufacturing process with minimum overhead on reconfiguration and quick optimization based on application scenarios. Additive manufacturing enables the creation of objects by successive addition of materials layer by layer. Proper utilization of the technology in the additive manufacturing process with reduced cost factors will be a better solution for creating quick cost-effective models in smart manufacturing.  



Moreover, with the roots of Industry 4.0 originated from Germany, many countries have developed certain industrial standards for smart manufacturing using information communication technology (ICT). Li \textit{et al.}\cite{li_smart_2018} analyzed the smart manufacturing architectures of different countries and developed a reference model for smart manufacturing.  





\subsubsection{AI in Robotics for Industries}

In modern industries, AI and robotics have emerged as a powerful combination for automating several tasks. According to Kipper \textit{et al.}\cite{kipper_scopus_2020}, different prospects of Industry 4.0 and beyond its capabilities are possible with an emphasis on robotic systems in smart industries. The literature already depicts a particular emphasis and trend of developing specialized robotics for different tasks in industries. For instance, in\cite{strozzi_literature_2017}, the applicability and importance of robots in production processes in smart industries are analyzed and discussed in detail. Robots are also used for ubiquitous manufacturing processes in smart industries\cite{wang2018comprehensive}.  

There are several applications of robotics in Industry 4.0 where AI is deployed, enabling robots to perform different tasks more efficiently. In the following, we highlight some of the key applications of Industry 4.0 where AI is deployed in robotics:

\begin{itemize}
    \item \textit{Assembling and Manufacturing }: AI has been proved very effective in enhancing robots' performance in assembling and manufacturing tasks. AI coupled with other advanced enabling technologies, such as vision systems, allows more flexibility by enabling robots to define, learn, and optimize processes in complex manufacturing sectors, such as aerospace. In such cases, cognitive robotic systems manage robot automation through Natural Language Processing (NLP), improved data handling using AI and ML approaches\cite{hu_irobot-factory_2019}. For instance, Hu \textit{et al.}\cite{hu_irobot-factory_2019} rely on AI in the implementation of an intelligent production line with robots in a smart factory using cognitive manufacturing for chip assembly.  
    \item \textit{Packaging}: In smart industries, robots are used for packaging products. AI can also help to further enhance robots' capabilities in packaging tasks, which could result in a quicker, accurate, and low-cost solution. For instance, Chen \textit{et al.}\cite{chen_intelligent_2019} proposed a Particle Swarm Optimization (PSO) based framework for time-optimal trajectory planning of robots in an AI-enabled intelligent packaging. 
    
    \item \textit{Production Task Scheduling by Robots}: Traditional assembly lines in industries have fixed tooling and fixed robots for the production process. Enhancement in the production process in modern industries can be carried out with the usage of re-configurable tools and mobile robots programmed with inherent task scheduling mechanisms and intelligent control units with the help of AI. Parente \textit{et al.}\cite{parente_production_2020} survey the challenges in Industry 4.0 towards holistic scheduling of tasks in the industries and the corresponding solutions for human-robot collaboration and appropriate decision-making tasks. In literature, several interesting AI-based solutions have been proposed for an intelligent task scheduling mechanism by robots. For instance, in\cite{messaoud_online_2020}, based on the demand for resources in the manufacturing process, using a slicing paradigm and AI, resources are allotted for robotic systems in Industry 4.0 environment. In\cite{ham_transfer-robot_2020}, simultaneous scheduling of tasks in the production sector for mobile robotic systems is carried out in a job shop environment proving flexible transfer of materials, providing an optimal benchmark performance.
    \textit{Customer Service}: AI-enabled robots are also used to entertain customers' queries about products or services, such as retail, in smart industries. Thanks to other enabling technologies, such as NLP, Computer Vision (CV), AI-based robots can interact with customers more humanly. 
\end{itemize}

\subsubsection{Generative Design}
Thanks to AI and other relevant technologies, such as vision and VR, the generative design delivers designs of products that we might not even have imagined or visualized. It helps to provide numerous options during the design phase in smart industries for a single drawing. All the designs developed using AI techniques are capable of meeting the specific goals set by the designers. From those pools of options, we can then select a design meeting the success criteria and expectations of the designers. The generative design technology harnesses massive computation power, and helps in creating forms with precise amounts of materials, only when required by ensuring maximum performance. In Generative design, the prototypes can be scanned and equipped with sensors providing real-time performance, which can be looped back into the design process.   

There are already some efforts in this direction. For instance, Yuan \textit{et al.}\cite{yuan2020attribute} explored the implications of Generative Adversarial Networks (GANs) based DL models in attribute-aware generative design for generating design with desired visual attributes. The authors also analyzed how GANs can help industries to meet the growing diversity in customers' needs.

\subsubsection{Market Adaption/Supply Chain}
Valuable supply chain integration for business planning, design, and implementation in modern industries largely depends on Industrial AI\cite{zhang_reference_2019}. Most of the supply chain paradigms from the perspective of Industry 4.0 are defined under one of the categories among Green, Resilience, Agile, and Lean\cite{ramirez-pena_achieving_2020}. In this regard, several efforts have been made. For instance, productive increase in small-scale industries also focused on urban production\cite{matt_urban_2020}, with a sustainable production process by implementing Industry 4.0 in their industrial supply chain. Through IIoT, automation, and predictive analytics, the maturity level of Industry 4.0 is analyzed through interviews and group discussion among focus groups\cite{tiwari_sustainability_2020}. Besides, accounting towards sustainability levels in the organization are reported along with the future maturity prospects of Industry 4.0. 

Moreover, Inventory management is analyzed from different perspectives of the customer to business models, combined to form an overall framework for modern industries\cite{zhang_state---art_2019}. Recommendation system for rescheduling in IIoT edge devices, particularly in low power wireless personal area networks enhances the performance in the network\cite{van_der_lee_interference_2019}. Here, the scheduling is performed using interference graphs and improved performance is observed compared to the rescheduling performed randomly. Scheduling jobs in the production line needs a collaborative approach and management of a cluster of resources efficiently. Xia \textit{et al.}\cite{xia_resource_2020} developed a collaborative task scheduling strategy for delay insensitive applications. 

\subsection{AI Techniques in Industry 4.0}
As discussed earlier, AI has been proven very effective in different applications of Industry 4.0. It not only enables predictive quality to ensure quality products but also helps in the optimization of different industrial operations. To fully explore the potential of AI/ML in industries, several interesting solutions including supervised, unsupervised, semi-supervised, and reinforcement learning techniques, have been proposed. In this section, we provide a detailed overview of AI and ML techniques deployed in Industry 4.0. For the sake of simplicity, we will divide it into two categories, namely (i) statistical/traditional methods, and (ii) DL techniques. 

\subsubsection{Statistical/Traditional AI/ML}
A vast majority of the efforts for AI-based solutions in Industry 4.0 rely on statistical/traditional techniques\cite{dalzochio_machine_2020}. For instance, Penumuru \textit{et al.}\cite{penumuru_identification_2019} rely on Computer Vision (CV) and traditional ML methods to embed industrial machines and robots with capabilities of automatically differentiating in four different types of materials including aluminum, copper, medium density fiberboard, and mild steel. To this aim, several classification algorithms including Support vector machine (SVMs), Decision trees, Random forests (RF), Logistic regression, and k-Nearest Neighbor (KNN) are trained on color features extracted from the materials. The experimental results showcase the validity of the algorithms in terms of their accuracy of classification. Logistic Regression and RF classifiers are also employed in\cite{candanedo2018machine} for automatic monitoring and analyzing the performance of an HVAC air conditioning system. The statistical ML techniques are also employed in\cite{romeo2020machine}, where several classification algorithms, such as RF, SVMS, and Logistic Regressions, are used for the estimation of machine specification data (i.e., machine geometry and design). 

Machines involved in continuous operations in industries are mostly prone to some unexpected faults, and the subsequent resuming of tasks after carrying out necessary countermeasures is often time-consuming. Some efforts on fault detection in smart industries also rely on statistical ML techniques. For instance, Angelopoulos \textit{et al.}\cite{angelopoulos_tackling_2020} performed a survey on the detection of faults by appropriate prediction using ML approaches and providing equivalent countermeasures in industry 4.0 applications. Moreover, the authors also addressed the ML-based interactions in the manufacturing process that occurs between machines and human operators by providing the platform for humans to be an in-the-loop technique with reduced errors.  

Down failure of machines intimation and its availability of carrying out subsequent manufacturing tasks are recommended features of predictive maintenance in smart industries. Data acquired from IIoT devices engaged in event-based monitoring of industrial machines are utilized for training ML models. Literature also suggests a tendency towards the statistical/traditional ML techniques, such as SVMs, RF, gradient boosting (GB), and extreme gradient boosting approaches, for predicting the failure probabilities of the machine\cite{calabrese_sophia_2020}. Similarly, enhancement in production planning and control tasks is also enabled using ML techniques with modern tools, data sources, and approaches. For instance, Cadavid \textit{et al.}\cite{cadavid_machine_2020} provide a detailed overview of state-of-the-art techniques involved in IIoT devices to collect data and analyze them using ML methods, including traditional ML and DL algorithms, used for different tasks involved in intelligent manufacturing processes. Bajic \textit{et al.}\cite{bajic_machine_2018} also evaluated the performance of different ML algorithms, such as SVM, KNN, and neural networks, in smart manufacturing by generalizing them for different tasks to provide solutions for enhancing the decision making challenges existing in industry 4.0 applications.  

In smart industries, embedded sensors placed in production lines acquire data regarding the quality and quantity of tasks performed during manufacturing and assembling processes. Traditional ML also contributes to task scheduling and optimization of the scheduling process to improve the efficiency of industrial machines. The traditional ML techniques are also employed for industrial prognosis, which deals with the estimation of the lifetime of machines used in industries due to the aging of machines and extended hours of utilization. In\cite{diez-olivan_data_2019} an overview of different ML techniques along with data fusion techniques for implementation of prescriptive, predictive, and descriptive prognostic models for industry 4.0 applications is provided.

Table \ref{tab:DLInd40} provides a summary of different works relying on statistical/traditional ML techniques for different applications of Industry 4.0. 

\begin{table*}[!hbtp]
  \begin{center}
    \caption{Contribution of ML algorithms for Industry 4.0 applications.}
    \label{tab:MLInd40}
    \begin{tabular}{|p{3cm}|p{4cm}|p{4cm}|p{5cm}|}
    \hline
    \textbf{References} & 
    \textbf{ML approaches} & 
    \textbf{Industry 4.0 applications} & 
    \textbf{Key Inferences}\\
    \hline
	Penumuru \textit{et al.}\cite{penumuru_identification_2019} & SVM, DT, RF, LR, KNN & Material classification & Surface classification of materials Aluminium, Copper, mild steel and fibreboard are done based on the RGB color model. \\    \hline
	Dalzochio \textit{et al.}\cite{dalzochio_machine_2020} & ML algorithms review & Predictive maintenance  & Using three research questions, the use-cases and implementation aspects of predictive maintenance are carried out.\\    \hline
	Angelopoulos \textit{et al.}\cite{angelopoulos_tackling_2020} & ML-based human–machine interactions & Tackling faults & Fault prediction, detection, and classification with the support of human-in-the-loop techniques. \\    \hline
	Odonovan \textit{et al.}\cite{odonovan_comparison_2019} & Embedded ML approaches & Cyber-physical interfaces & Fog computing provides reliable services with secure and consistent support for industrial applications. \\    \hline
	Calabrese \textit{et al.}\cite{calabrese_sophia_2020} & Event-based IIoT and ML approaches & Predictive maintenance & Data from the IIoT devices are utilized for effective failure prediction of the smart machines.\\    \hline
	Khayyam \textit{et al.}\cite{khayyam_novel_2020} & Hybrid ML algorithm & Big Data modeling & Over-fitting issues in the industrial Big Data are effectively handled using the hybrid technique.\\    \hline
	Cadavid \textit{et al.}\cite{cadavid_machine_2020} & ML-PPC & Production planning and control & The techniques are used to adapt the system towards changes in the manufacturing sector through modern tools and data sources. \\    \hline
	Bajic \textit{et al.}\cite{bajic_machine_2018} & SVM, KNN & Smart manufacturing & Enhanced decision making skills are driven in smart manufacturing for generalized tasks. \\    \hline
	Diez-olivan \textit{et al.}\cite{diez-olivan_data_2019} & ML and data fusion & Industrial prognosis & Prognostic models are used to predict the status of machines considering its wear and tear conditions. \\    \hline
	Mohanta \textit{et al.}\cite{mohanta_management_2020} & ML approaches & Risk management & Data and resources are managed effectively by addressing the issues related to uncertainty, volatility, ambiguity and complexity. \\    \hline
	Shukla \textit{et al.}\cite{shukla_industry_2018} & ML and NLP  & Aeronautical applications & CPS works in conjugation with the ML models for better interactions between machines.  \\    \hline
	Li \textit{et al.}\cite{li2020machine} & Hybrid metaheuristic & Production rescheduling  & Tailored delay and frequencies through ML models drives efficient job scheduling. \\    \hline
    \end{tabular}
  \end{center}
\end{table*}
\textbf{}

\subsubsection{Deep Learning}
Similar to other domains, DL algorithms have also been widely explored for different applications of Industry 4.0. The literature shows some interesting DL-based solutions for different tasks in Industry 4.0. For instance, Miskuf \textit{et al.}\cite{mivskuf2016comparison} proposed a DL-based approach to data analytics tasks in Industry 4.0 applications, such as machine failure prediction and recommendation systems. An open-source framework with H2O ML is used for implementing the DL model, which is evaluated on a text recognition data set obtained from the UCI repository. The authors also provide a comparison of the DL method with traditional ML techniques, such as logistic regression.

In\cite{villalba2019characterization}, a DL-based framework, relying on a geometric DL architecture, has been proposed for lean management of cyber-physical networks. In\cite{villalba2019characterization}, EEG sensors are configured in conjugation with DL frameworks to assess the behavioral patterns in lean management. The characterization involves the usage of soft sensors for the study of the acquired signals for solving the behavioral patterns. However, this framework is effectively deployed by authorized heads of smart industries for planning sustainable organizational goals and providing reliable smart manufacturing systems. DL-based soft sensors are also used for lean shop floor management by observing the electrical activities of the human brain using EEG sensors\cite{schmidt2020industry}. The analysis is used to provide effective correlations from the recorded data on the brain’s activity for assessing the shop floor management and provides better guidance for the decision-makers of smart industries that implements industry 4.0.  

DL techniques have also been employed in other several applications of Industry 4.0. For instance, Subakti \textit{et al.}\cite{subakti2018indoor} proposed a DL-based framework for automatic detection of machines and their parts as a key component/module of their Augmented Reality (AR) based solution for visualization and analysis of machines and interaction with them in an indoor environment. To this aim, an existing deep model namely MobileNet\cite{qin2018fd} has been fine-tuned on machine images. The selection of MobileNet is motivated by its architecture being suited for devices with less computational power. Pokhrel \textit{et al.}\cite{pokhrel2020multipath}, on the other hand, proposed a Deep Q-Network based framework for the implementation of multi-path communication in an industry 4.0 ecosystem. The authors also analyzed the challenges rooted due to the convergence of information and operational technologies with proof of concepts by carefully addressing the trade-offs in their proposed scheme. In\cite{cao2020multi}, task offloading and multi-channel access of mobile edge computing operations in industry 4.0 is carried out using a multi-agent deep reinforcement learning approach. In the developed scheme, the success rate of the channels is highly improved with a significant reduction in computation delay as the edge devices involved in the operations are cooperative.  

DL approaches are also effectively employed for the digital transformation of the construction industry, allowing to carefully and efficiently manage the complex environments through innovations established among the educational sectors, public, and government\cite{kraus2019construction}. Table \ref{tab:DLInd40} provides a summary of some key works relying on DL methods for different tasks in Industry 4.0.  

\begin{table*}[!hbtp]
  \begin{center}
    \caption{Contribution of DL algorithms for Industry 4.0 applications.}
    \label{tab:DLInd40}
    \begin{tabular}{|p{3cm}|p{4cm}|p{4cm}|p{5cm}|}
    \hline
    \textbf{References} & 
    \textbf{DL approaches} & 
    \textbf{Industry 4.0 applications} & 
    \textbf{Key Inferences}\\
    \hline
	Mivskuf \textit{et al.}\cite{mivskuf2016comparison} & H2O framework & Multi-class vs DL & Data analytics and cloud computing provides foundations for industry 4.0 applications\\    \hline
	Villalba \textit{et al.}\cite{villalba2019characterization} & DL sensor & Lean management  & EEG sensors are used for assessing the behavioural patterns and it aids in sustainable smart manufacturing systems.\\    \hline
	Pokhrel	\textit{et al.}\cite{pokhrel2020multipath} & Deep Q-Network & Multipath communication framework & Information and operational technologies are integrated for implementation of flexible manufacturing in industry 4.0 ecosystem.\\    \hline
	Schmidt \textit{et al.}\cite{schmidt2020industry} & DL sensor  & Shopfloor management & Brain’s activity are correlated for effective decision making in smart industries\\    \hline
	Cao \textit{et al.}\cite{cao2020multi} & Multi-agent DRL  & Mobile edge computing & Cooperative operation of edge devices significantly reduces the delay and improves the performance of the channels. \\ \hline
	Kraus \textit{et al.}\cite{kraus2019construction}  & DL  & Construction enterprises & Support for digital transformation through innovations\\ \hline
	Villalba \textit{et al.}\cite{villalba2019deep}  & DNN sensor  & Printing Industry & Automated quality control with classification accuracy of 98.4\%\\ \hline
	Elsisi \textit{et al.}\cite{elsisi2021deep}  & YOLOv3 & Energy management & Optimizes energy is smart buildings by detecting the number of persons.\\ \hline
	Ma \textit{et al.}\cite{ma2019research}  & Multi-source input neural network  & Sea clutter processing & Sea clutter reflectively prediction through remote sensing provides enhanced accuracy with average prediction error of 1.82 dB.\\ \hline
	Francis \textit{et al.}\cite{francis2019deep}  & CNN &  Additive manufacturing & Distortion prediction and geometric inaccuracies in laser-based fabrication process \\ \hline
	Demertzis \textit{et al.}\cite{demertzis2020anomaly}  & Deep Autoencoder & Anomaly detection & Smart contracts in Industry 4.0 are secured is association with blockchained network. \\ \hline
	Essien \textit{et al.}\cite{essien2020deep}  & LSTM,   Autoencoders & Smart Manufacturing & Data from a metal packaging plant are used for  empirical analyses and machine speed prediction. \\ \hline
	Leng \textit{et al.}\cite{leng2021loosely} & DRL & PCB manufacturing & Sustainable and environment friendly decision making on orders are ensured. \\ \hline
    \end{tabular}
  \end{center}
\end{table*}
\subsection{Datasets for Industry 4.0}
We envision that datasets play a crucial role in all categories of learning frameworks in AI-based industrial applications, where it is one of the vital ingredients to carry out data analytics. The feasibility of an ML algorithm in an application is also constrained by the availability of a sufficient amount of training data. To this aim, in this section, we provide a detailed description of some of the benchmark datasets available for the training and evaluation of AI frameworks for Industry 4.0 applications. 

\subsubsection{Production datasets}
Sustainable and reliable manufacturing processes in the production sectors mostly demand precision in decision-making through the usage of AI algorithms trained on appropriate production datasets. Based on the category of industries, and the products being manufactured these datasets play a significant role in estimating the prospects of those industries. Honti \textit{et al.}\cite{honti2020data} collected and used a dataset, namely the I4.0+ dataset, for evaluating the economic and innovation index with the support of NLP and data mining principles. In addition, authors in\cite{hahn2020industry}, used 200+ use cases of innovation in the supply chain for customizing the on-demand standard services for customers. In another work by Romeo \textit{et al.}\cite{romeo2020machine}, using two real-world datasets, the design support systems in industries are validated using various DL and traditional ML approaches. Similarly, Elhoone \textit{et al.}\cite{elhoone2020cyber} use a parametric design dataset for additive manufacturing applications using ANN frameworks. This lays a foundation for carrying out cyber additive manufacturing processes. In the case of Oil and Gas industry applications, work with massive seismic datasets was used for evaluating the uncertainties using Type-2 fuzzy sets.

\subsubsection{Cyber-Physical Systems datasets}
The core security-centric operations in Industry 4.0 eco-system could be strengthened with the AI systems monitoring the functional aspects of the industry through the aid of CPS. Data accumulated from CPS could be highly beneficial to train AI frameworks in imparting high-end security solutions. For instance, CPS datasets are used by Moustafa \textit{et al.}\cite{moustafa2018new}, with the aid of a mixture-hidden Markov model for addressing the threat issues in industrial systems. It also ensures an intelligent level of safeguarding the machines from anomalous activities. Pan \textit{et al.} \cite{pan2015developing,CPS1} proposed and collected a CPS dataset for training and evaluation of intrusion detection techniques for power systems. The original dataset is composed of 15 subsets each covering 37 different power system events/scenarios including 8 natural events, 28 attacks, and 1 non-event. Moreover, the dataset supports binary, three class, and multi-class classification tasks. A similar type of dataset is presented in \cite{morris2014industrial,morris2011control}, where samples of cyber attacks against two different types of industrial control systems, including a gas pipeline and water storage tank, are provided. The dataset is intended to help researchers in training and evaluation of ML algorithms for differentiating in attacks and non-attacks/normal network behavior. Another dataset from the same authors is presented in \cite{morris2015industrial}, where the authors proposed a dataset for industrial control system simulation and data logging on intrusion detection. The dataset provides six different types of features including \textit{modbus frame}, \textit{label}, \textit{type of specific attacks}, \textit{source}, \textit{destination}, and \textit{time stamp}.  
In \cite{beaver2013evaluation}, on the other hand, a dataset is collected for training and evaluation of ML frameworks for detection of malicious Supervisory Control and Data Acquisition (SCADA) Communications in gas pipelines. The dataset provides 7 different types of the data injection attack, namely \textit{negative values}, \textit{burst values}, \textit{fast change}, \textit{slow change}, \textit{single data injection}, \textit{value wave}, and \textit{set point value injection}, each aiming to manipulate the system's response in a different way.

\subsubsection{Image datasets}
The literature also provides several image datasets for training and evaluation of AI models in different applications of Industry 4.0. For instance, Penumuru \textit{et al.} \cite{penumuru_identification_2019} proposed and collected an image dataset to train and evaluate AI models for the identification and classification of different types of materials in the context of industry 4.0. The dataset mainly covers images from four different types of material, namely Aluminium, Copper, Medium density fibreboard, and Mild steel. In \cite{luo2019benchmark}, a benchmark image dataset for industrial tools has been collected. The dataset is composed of 11,000 images from eight different categories of 24 industrial tools, which are manually analyzed and annotated by experienced workers. All the images are acquired through a Kinect 2.0 sensor placed at a distance of 1 to 5 meters, and a resolution $1024 X 575$ pixels and $512 X 424$ depth frames. In \cite{KAGGLE_DATASET}, an image dataset namely Severstal: Steel Defect Detection has been proposed for training and evaluation of image-based solution for defect detection in steel. The dataset provides a total of 18,106 steel images, divided into training and test sets containing 12,600 and 5,506 images, respectively. All the images are annotated with either an image with (i) no defects, (ii) a defect of a single class, or (iii) defects of multiple classes. 

Automating the product inspection in smart industries, which is one of the key applications of Industry 4.0, can be driven through the usage of AI algorithms trained on an appropriate dataset. Schwebig \textit{et al.}\cite{schwebig2020compilation} reported one such instance for automating the solder joint inspection of Printed Circuit Board (PCB) assembling. The dataset is composed of a large number of images taken through optical inspection systems from different manufacturers. Moreover, the dataset contains both color and gray-scale images. Ntavelis \textit{et al.}\cite{ntavelis2020same}, on the other hand, proposed a GANs-based framework for generating/augmentation of industrial images utilizing a smaller set of natural images. The synthesized datasets can then be utilized for training DL algorithms. The method is evaluated on a dataset of pump parts, which is composed of 3 channel images. Moreover, the images are acquired under different sources of illumination.  

\subsubsection{Text datasets}
In the modern world, businesses rely on AI tools to automatically analyze and extract people's sentiment towards their products, businesses, or a service \cite{hassan2019sentiment,hassan2020visual}. In smart industries, sentiment analysis can help businesses to improve the quality of their products by analyzing customers' responses to their products. There are already some efforts in this regard. For instance, in \cite{ireland2018application}, sentiment analysis of customers' feedback on different products has been carried out to analyze the drawbacks of the product, and improve the products' quality accordingly. To this aim, a dataset composed of customers' reviews on a product from the e-commerce website Amazon has been used to evaluate the performance of the proposed sentiment analysis framework. A similar type of dataset is collected in \cite{syamala2020filter}, where sentiment analysis has been carried out on a large collection of customers' reviews on Amazone products. A customers' reviews dataset,composed of 26,993 positive and 13,203 negative reviews, is also collected in \cite{pasmawati2020exploiting}.  The positive and negative reviews are then processed by some NLP techniques to extract and classify product design attributes. In \cite{yilmaz2017social}, on the other hand, a Twitter has been crawled to analyze people's response to Industry 4.0. To this aim, a dataset composed of Industry 4.0 related tweets is collected and used for assessing and classification of sentiments' polarity towards Industry 4.0 into \textit{positive}, \textit{negative}, and \textit{neutral}. The analysis was carried out using text mining to assess the impact of the words related to Industry 4.0. 

Besides sentiment analysis, text-based datasets are also used for other applications/tasks in Industry 4.0. For instance,  In \cite{pejic2020text}, a text mining dataset, composed of Industry 4.0 job advertisements, has been proposed and collected. To collect the dataset, different online sources/websites of job advertisements are crawled. One of the main motivations of the dataset is to collect relevant information on the knowledge and skills required for industries. 

\subsubsection{Survey datasets}
In smart industries, surveys are key to extract meaningful insights from the data collected from the workgroups in smart industries through means of feedback and surveys. Several such datasets have been used in different applications of Industry 4.0. For instance, Frank \textit{et al.}\cite{frank2019industry} collected a survey dataset from 92 different industries and assessed the implementation patterns of smart industries through cluster analysis. The dataset is used for evaluating the systematic adoption of base supportive technologies for modern industries. In another work by Medi{\'c} \textit{et al.}\cite{medic2019hybrid}, the authors collected a European manufacturing survey dataset for assessing the technology for carrying out strategic decisions through the usage of Fuzzy based analytics.

\begin{table*}[!hbtp]
  \begin{center}
    \caption{A summary of some of the benchmark datasets for training and evaluation of AI techniques in Industry 4.0 applications.}
    \label{tab:DatasetInd40}
    \begin{tabular}{|p{2cm}|p{2cm}|p{2cm}|p{2cm}|p{2cm}|p{5cm}|}
    \hline
    \textbf{Dataset categories} &
    \textbf{References} & 
	\textbf{Datasets} & 
	\textbf{Industry 4.0 applications} & 
	\textbf{AI framework} & 
	\textbf{Key Inferences}\\ \hline
    \multirow{5}{*}{\parbox{2cm}{Production datasets}}
	& Honti \textit{et al.}\cite{honti2020data} & I4.0+ & Industry 4.0 readiness measures & NLP and data mining & Economic and innovation index are evaluated \\    \cline{2-6}
	& Hahn \textit{et al.}\cite{hahn2020industry} & 200 + company use cases & Supply chain innovation & Structured content analysis & Customized and on demand standard services for customers are suggested.  \\  \cline{2-6}
	& Romeo \textit{et al.}\cite{romeo2020machine} & 2 real use datasets & Design Support System & Different ML and DL approaches & Validation of support system are assessed  by comparing accuracy, speed and \\  \cline{2-6}
    & Elhoone \textit{et al.}\cite{elhoone2020cyber} & Parametric design & Additive manufacturing & ANN & Lays foundation for cyber additive manufacturing by dynamically alloting digital resources.\\ \cline{2-6}
    & Villalonga \textit{et al.}\cite{villalonga2020cloud} & On-production & Data driven reasoning & Reinforcement learning & Surface roughness quality is ensured in the milling process. \\   \hline	

	\multirow{7}{*}{\parbox{2cm}{CPS datasets}} & Moustafa \textit{et al.}\cite{moustafa2018new} 
	& CPS dataset and UNSW-NB15 & Intelligent safeguarding of industrial systems & Mixture-hidden markov model & The threat intelligence scheme evaluates the anomalous activities in industrial systems.\\ \cline{2-6}
    & Pan \textit{et al.}\cite{pan2015developing} & Power system events & Intrusion detection systems & Path mining & Disturbances, control and cyber attacks are classified.  \\     \cline{2-6}	
	& Morris \textit{et al.}\cite{morris2014industrial} & Network traffic & Cyber attacks in SCADA systems & ML techniques & Traffic attacks on the SCADA systems are analyzed \\     \cline{2-6}	
	& Morris \textit{et al.}\cite{morris2015industrial} & Industrial control system & Intrusion Detection System & ML techniques & Gas pipeline network data log was trained and tested. \\     \cline{2-6}	
	& Beaver \textit{et al.}\cite{beaver2013evaluation} & Remote Terminal Unit communications & Malicious SCADA Communications & ML techniques & Normal and data injection attacks are assessed under various scenarios.\\    \cline{2-6} 
	& Li \textit{et al.}\cite{li2017intelligent} & CPS dataset & Predictive maintenance & ANN & Fault diagnosis and prognosis are carried out for systematic maintenance of smart machines.\\  \hline	

	\multirow{7}{*}{\parbox{2cm}{Image datasets}} 
	& Penumuru \textit{et al.}\cite{penumuru_identification_2019} & Industrial materials & Industrial materials & ML techniques  &  For various camera orientations, illumination and focal length of image data, the robustness of the techniques are evaluated.\\     \cline{2-6}	
	& Luo \textit{et al.}\cite{luo2019benchmark} & Industrial tools & Benchmark standards & Mean averaging & Provides ground truth images for evaluating industrial systems.\\     \cline{2-6}
	& Kaggle \cite{KAGGLE_DATASET} & Steel & Defect identification & ML techniques  & Helps to localize and classify surface defects on steel sheets.\\     \cline{2-6}	
	& Schwebig \textit{et al.}\cite{schwebig2020compilation} & Images of PCBs  & Automatic optical inspection & CNN & Automated inspection of solder joints in assembling are assessed without human interference\\ \cline{2-6}
    & Ntavelis \textit{et al.}\cite{ntavelis2020same} & Industrial images & Augmentation of industrial images & GAN & With handful of industrial images, augmentation is carried out using DL.\\     \cline{2-6}
    & Ozdaugouglu \textit{et al.}\cite{ozdaugouglu2020predictive} & Bibliometric dataset & Text classification & Fuzzy based multi-criteria technique & The computing techniques reveals the mainstream research direction towards Industry 4.0 \\     \cline{2-6}
    & Ntavelis \textit{et al.}\cite{ntavelis2020same} &  Industrial images & Augmentation of industrial images & GAN & With handful of industrial images, augmentation is carried out using DL.\\    \cline{2-6}
	& Nguyen \textit{et al.}\cite{nguyen2020systematic} & Massive seismic & Oil and Gas industry & Type-2 fuzzy sets & Evaluates various aspects of using Big Data with their benefits and uncertainties involved. \\ \hline
    
    \multirow{7}{*}{\parbox{2cm}{Text datasets}} 
    & Hassan \textit{et al.}\cite{hassan2019sentiment} & Customers' responses & Sentimental analysis & Text mining & Social media feedback helps to estimate people's emotions, feelings and opinions.\\     \cline{2-6}
	& Ireland \textit{et al.}\cite{ireland2018application} & Customers' reviews on Amazon products & Sentimental analysis & Statistical approach & Suggests creative and logical design aspects on the products. \\     \cline{2-6}	
	& Syamala \textit{et al.}\cite{syamala2020filter} & Customers' reviews on Amazon products & E-commerce & Decision Tree & Decision making towards better product recommendation.\\     \cline{2-6}	
	& Pasmawati \textit{et al.}\cite{pasmawati2020exploiting} & Product reviews & Customer online reviews & ML techniques & User experience are evaluated for providing better business strategies and product design. \\     \cline{2-6}	
    & Yilmaz \textit{et al.}\cite{yilmaz2017social} & Tweets dataset & Sentimental analysis & Text mining &  Positive, Negative and Neutral aspects of words related to Industry 4.0 is analyzed.\\     \cline{2-6}
	& Pejic \textit{et al.}\cite{pejic2020text} & Job advertisements & Knowledge requirement for smart industries & Text mining & Helps to chose aspiring and potential candidates for Industry 4.0 organizations.\\    \hline	
	
	\multirow{2}{*}{\parbox{2cm}{Survey datasets}} 
	& Frank \textit{et al.}\cite{frank2019industry} & Survey dataset from 92 companies & Implementation patterns & cluster analysis & Systematic adoption of base supporting technologies are evaluated. \\  \cline{2-6}
	& Medi{\'c} \textit{et al.}\cite{medic2019hybrid} & European	manufacturing survey dataset & Technology assessment in manufacturing & Fuzzy analytic hierarchy & Assists manufactures to provide strategic decision making. \\  \hline	
	\end{tabular}
  \end{center}
\end{table*}
\section{Big Data Analytics in Industry 4.0}
\label{bigdata}

\begin{figure*}[!t]
\centering
\centerline{\includegraphics[height=7cm, width=14cm]{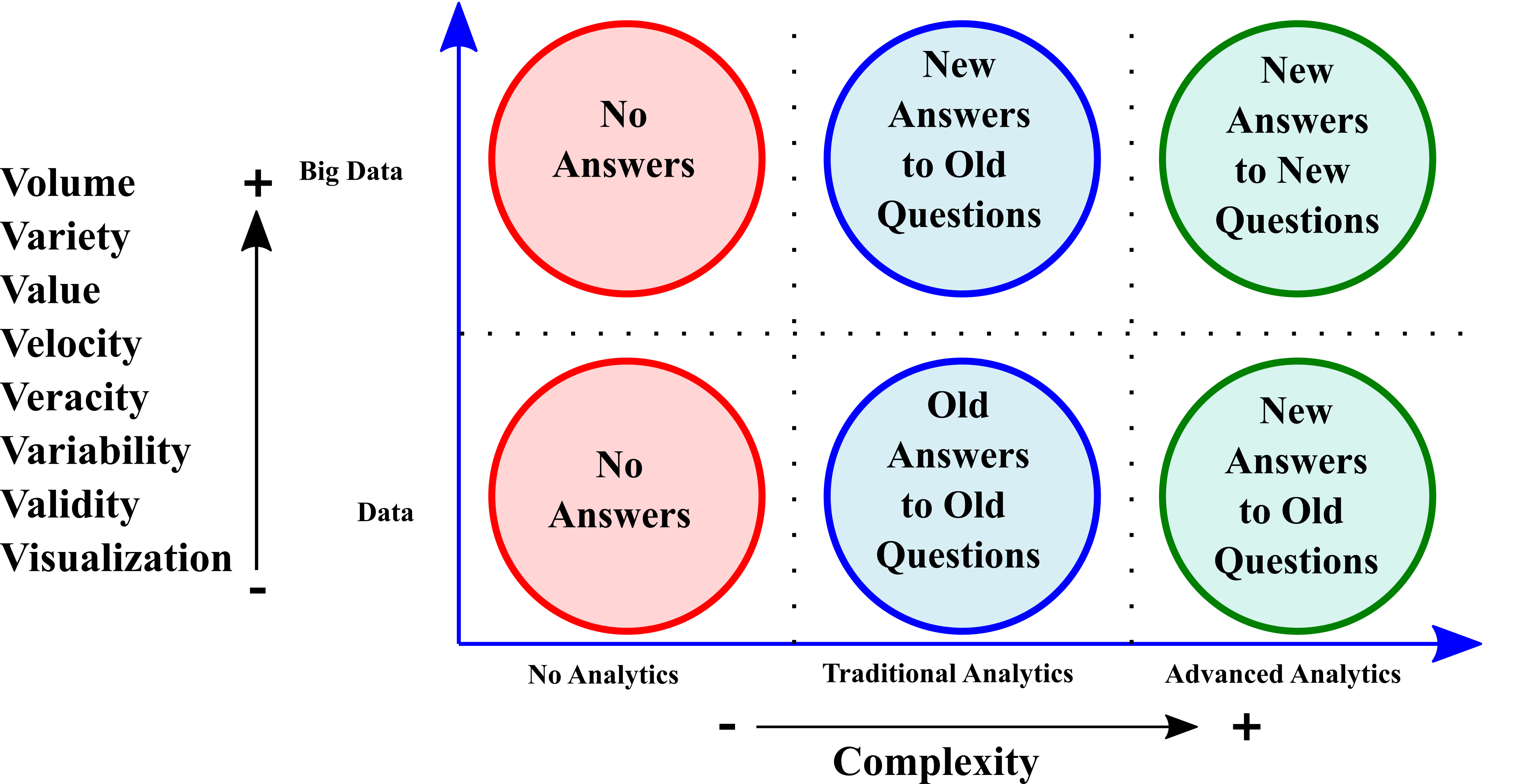}}
\caption{Answers to the questions based on the analytics on Industrial Big data.}
\label{fig:bigdata}
\end{figure*}

Big Data is one of the ten technologies driving the smart industries\cite{aceto_survey_2019}. More recently, several works emphasized the importance of Big Data methods for extracting meaningful insights on different aspects of smart industries. For instance, Cui \textit{et al.}\cite{cui_manufacturing_2020} provide a detailed analysis of how Big Data can help in different applications of smart industries. The authors also highlight the usage of appropriate hardware, sharing of resources, integration of subsystems, data prediction, and sustainability as core enablers for Big Data applications. Similarly, Belhadi \textit{et al.}\cite{belhadi_understanding_2019} analyzed the strategic implications of Big Data in smart manufacturing processes by showcasing several use-cases. The Big Data solutions in Industry 4.0 are centralized providing new answers to new questions as visually depicted in Figure \ref{fig:bigdata} along with the 8V's of Big Data. These 8V's used in literature to define Big Data include Volume, Variety, Value, Velocity, Veracity, Variability, validity, and visualization. Moreover, the complexity of the methods is directly proportional to the level of analytics, where the complexity of the methods increases as we go for advanced analytics to find newer and detailed answers.   

In the next subsections, we discuss the applications and most commonly used Big Data techniques in Industry 4.0 applications. 

\subsection{Big Data Applications in Industry 4.0}

\subsubsection{Big Data in Smart Manufacturing}
Sustainability in smart manufacturing is supported when the smart machines and robotic systems employed in industrial sectors learn from data. Big Data and the real-time analytics performed on the data are widely used for ubiquitous manufacturing\cite{wang2018comprehensive, ren_comprehensive_2019}. To assess and evaluate Big Data solutions in smart manufacturing, Nagadi \textit{et al.}\cite{nagadi_hybrid_2018} proposed a hybrid simulation tool composed of several modeling techniques. The tool allows mapping characteristic data of different machines, modeling tools, and other captured plans into an object-oriented simulation. In\cite{gupta_big_2020}, Big Data applications in smart industries are assessed using a lean six sigma process for identifying the gaps and opportunities for development in the manufacturing sector and Big Data-related research for sustainable manufacturing processes.

Early fault detection and diagnosis in smart manufacturing is one of the key applications relying on Big Data analytics. To this aim, several interesting solutions have been proposed. For instance, Yu \textit{et al.}\cite{yu_global_2020} proposed a distributed PCA-based solution namely the DPCA model for fault detection and diagnosis in smart manufacturing. The simplicity and ease of implementation make it a suitable solution for real-time processing, which is one of the key requirements of Industry 4.0 applications.  In\cite{wang_integrated_2017}, Big Data is employed in a smart factory for candy packing where the semantic data and ontology knowledge are combined for enabling personalized packaging. 

\subsubsection{Additive manufacturing}
Additive manufacturing allows industries to develop low-cost personalized and on-demand products or complex parts of a machine in a short time, which results in low energy consumption and fewer waste materials. In additive manufacturing, cost prediction of a product (i.e., machine or a part of a machine) is an important factor that has a direct impact on the production and is a very challenging process. In smart industries, thanks to ML and Big Data techniques, the cost of different products can be accurately predicted. However, accurate predictions require a sufficient amount of manufacturing knowledge/information, which could be obtained from the raw data produced during the manufacturing and supply chain process. 

In literature, several interesting solutions have been proposed for analyzing different aspects of additive manufacturing. In\cite{bonnard_data_2019}, usage of digital data model thread for additive manufacturing is presented with the solutions for data management to meet the requirements of modern industries. In\cite{lee_design_2018}, smart logistics are implemented for Industry 4.0 using a warehouse management system driven by IoT devices employed for collecting real-time data. Majeed \textit{et al.}\cite{majeed2019framework} on the other hand propose a Big Data-based analytical framework to analyze and optimize the production performance of additive manufacturing processes. The evaluations of the proposed framework are carried out on real-life industrial use-cases and are found useful for all stakeholders. Francis \textit{et al.}\cite{francis2019deep} employed a DL-based framework for the analyses of industrial Big Data acquired through several sensors to deal with the inaccuracy of the fabricated parts due to distortion. The DL models are used to accurately predict the distortion in the fabricated parts. 

\subsubsection{Product/machine design}
Big Data along with AI and ML techniques are also contributing to product/machine designs\cite{li2015big}. In modern industries, the designing job is shifting towards AI and Big Data, and the role of the human designers could be reduced to directions/supervision only. The majority of the design aspects will be covered by AI with the help of Big Data techniques to extract key insights from historical data. For instance, AI algorithms can help the human designers in finding alternatives, risk assessment, and incorporating sensor-based data, generating design precursors, and in the optimization of supply-chain processes\cite{li2015big}. There are already some efforts in this direction. For instance, Yu \textit{et al.}\cite{yu2016product} propose a data-driven design for a product. The authors also analyze the connection between data and a product design, and how Big Data can help in the process. 

\subsubsection{Critical Analysis in Smart Industries }
Big Data enables industries to conduct a critical analysis of different aspects of a business. To this aim, various frameworks have been proposed in the literature. For instance, Rehman \textit{et al.}\cite{ur_rehman_role_2019} provide a detailed analysis of various analytical frameworks for Big Data processing with the help of appropriate use-cases. In\cite{raut_linking_2019}, a hybrid two-stage model supported by Artificial Neural Networks (ANN) is developed to explore sustainability in the performance of a business. Ren \textit{et al.}\cite{ren_comprehensive_2019} also found Big Data analytics very useful in sustainable smart manufacturing. Belhadi \textit{et al.}\cite{belhadi_understanding_2019} on the other hand applied Big Data analytics in multiple case studies on manufacturing during the manufacturing process in industries. 

Since the data collected from industrial appliances are highly prone to different types of attacks as detailed in Section \ref{sec:challenges}, Big Data analytics are also used to guard against different attacks on data, such as privacy and illegal access, and use to data. For instance, Keshk \textit{et al.}\cite{keshk_privacy-preserving_2018} analyzed the role of Big Data analytics in ensuring data privacy and security of sensitive data. To this aim, an independent component analysis (ICA) based framework has been proposed to transform raw Cyber-physical systems information into a new shape. The proposed technique is evaluated against several state-of-the-art methods on real Cyber-physical systems data, where significantly better results are observed for the proposed method. 

\subsubsection{Advanced Information Management in Industries}
Transformation towards smart manufacturing is driven by effective information management analyzed and learned from industrial data. Dynamic demands in the industries are learned from the data and analyzed with the aid of an autonomous model\cite{qu_smart_2019}. Management of information from the smart machines is performed through humans in the loop interactions carried out by learning from the industrial data and by the adoption of AI in the process\cite{turner_intelligent_2019}. Data regarding appropriate usage of the machines, resource sharing, sustainability predictions in industries are key information extracted from the real-time analytics of Big Data in smart industries\cite{cui_manufacturing_2020}. Moreover, the information gained from machines has transformed knowledge-based systems into smart and sustainable manufacturing systems and provides a platform for cyber-physical production processes\cite{vernadat_information_2018}.  

\subsubsection{Risk Analysis using Big Data} 

Risk management, such as reputational risk management, can enable the industrial sector to deploy its massive datasets in an efficient and optimized manner while extracting accurate and precise information. Moreover, real-time implementations can further improve risk management and render it more precise and timely in risk capturing, evaluation, and mitigation in addition to the identification of hidden values from the data via advanced measures of data quality monitoring. It is essential to note that such data measures can be useful for non-quantifiable risk typologies. While the complexity and dimensions of data collected by smart manufactures' sensors (from the huge amount of daily inputs in the manufactures' operating systems) are dramatically increasing, it presents a great opportunity for AI-based risk extraction and assessment while processing and extracting values efficiently and timely form Big Data. As a result, useful information for risk assessment can be returned in real time\cite{dicuonzo2019risk}.

Although the integration of Big Data in risk assessment can notably provide a vital advantage for real-time data quality monitoring in I 4.0, the real-time processing of quietly variable amount of collected data demands not only new mechanisms and tools, but also the broadening of mathematical and statistical information, and AI innovations, primarily oriented to both quantitative and qualitative analysis of data. As a consequence, data interpretation and transformation into a high added-value become notably more efficient which currently requires elaboration and design of new skills and knowledge for the transformation of data into advanced industrial strategic resources and simulations of industrial supply chain\cite{vieira_simulation_2019}.

\subsection{Big Data Methods in Industry 4.0}
Streaming Big Data in Industry 4.0 applications could be processed either at local devices, edge devices, or remote machines. Attributes of industrial Big Data, such as speed, bandwidth, access mechanisms, and storage aspects, play a crucial role in the decision-making of choice of computation resources. In addition, data analysis and visualization of the data help to gain meaningful insights out of them. Following are the most prominent analytical methods carried out for industrial Big Data.

\subsubsection{Descriptive Analytics} 
Descriptive analytics, which refers to the interpretation and analysis of historical data of business, allows businesses to better analyze and understand the changes and their causes over a while. Such analysis plays a critical role in the decision-making process by identifying the strength and weaknesses of a business or a particular strategy adopted and practiced over a while. There are two main data collection techniques, namely (i) data aggregation, and (ii) data mining, involved in collecting data for descriptive analysis\cite{araz2020role}. Since in the industries data is collected from several sources, before processing the data for meaningful insights, data is firstly collected, aggregated, and then analyzed to obtain manageable information. Moreover, the Majority of the business/financial metrics, such as year-over-year cost/growth changes, number of customers, etc, are the product of descriptive analysis\cite{araz2020role}. 

In smart industries, descriptive analytics is carried out regularly to see what went wrong and where the business excelled. The literature also suggests the importance of descriptive analytics in Industry 4.0 applications, and several interesting solutions are proposed to this aim. For instance, Ciano \textit{et al.}\cite{ciano2020one} used the industrial data to carry out descriptive analytics for improving the lean production techniques and integrating them with Industry 4.0 technologies. To this aim, the authors analyzed in several case studies.  Descriptive analytics also helps to provide smart process planning by characterizing the use of data and providing analysis on what happened in the industry\cite{araz2020role}.  

\subsubsection{Predictive Analytics}

Predictive analytics allows industries, businesses, and organizations to be proactive by predicting future events based on historical data. Future event detection in an industrial environment is very critical in several applications of Industry 4.0. Some key applications include predictive maintenance, product cost, lifetime, and failure prediction. Using statistical modeling, ML, and other sophisticated predictive analytic tools, industries could use the past and current data to reliably forecast trends and behaviors of machines. Moreover, it helps the industrial control group to predict the inventory requirements, manage shipping schedules and configure store layouts to maximize the productivity and sales of the products. 

The literature shows that predictive analytics has been widely deployed in Industry 4.0. For instance, Zonta \textit{et al.}\cite{zonta2020predictive} utilize predictive analytics for predictive maintenance in smart industries based on historical industrial Big Data. The authors also presented time-based schemes for addressing the challenges associated with predictive maintenance in the manufacturing process. In addition, the impact of Big Data is well addressed with a taxonomy of monitoring tasks in industries by carrying out predictive analytics.  Similarly, in\cite{nordal2020modeling}, a predictive maintenance management architecture is modeled by applying system engineering methods on existing maintenance management systems. In addition, with the support of predictive analytics, establishing AI-based maintenance management provides a reference architecture. In\cite{sahal2020big}, on the other hand, a wide range of applications of predictive analytics is explored in energy and transportation industries along with open source tools used for deployment. According to Tiwari \textit{et al.}\cite{tiwari2020sustainability}, high levels of maturity in smart industries could be achieved through predictive analytics and AI for achieving sustainability in industrial operations. 

\subsubsection{Prescriptive Analytics}
Prescriptive analytics allows industries, businesses, and organizations to make better decisions based on historical data. It is important to mention that prescriptive and predictive analysis are closely linked where the former makes use of the latter in defining a course of action based on predictions made by predictive analytics methods. 

Big Data analytics in modern industries is moving from predictive to prescriptive analytics\cite{groger2018building}. It not only predicts what will happen but also defines a course of actions that can be taken to benefit from these relations. Prescriptive analytics in industries could also express the impact of each alternative, and helps to avoid the problems by choosing alternate solutions. Prescriptive analytics in industries helps to achieve optimal decisions with adaptive and supports automation of tasks by driving them to operate in a time-constrained manner. According to Bousdeki \textit{et al.}\cite{bousdekis2020sensor} prescriptive analytics of time-dependent parameters makes the decision process much easier by providing reliable and realistic outcomes.

Table \ref{tab:BigdataInd40} summarizes some key works employing different methods of Big Data for Industry 4.0 applications. 
\begin{table*}[!hbtp]
  \begin{center}
    \caption{Contribution of Big Data approaches for Industry 4.0 applications.}
    \label{tab:BigdataInd40}
    \begin{tabular}{|p{3cm}|p{3cm}|p{4cm}|p{5.5cm}|}
    \hline
    \textbf{Big Data approaches} & 
    \textbf{References} & 
    \textbf{Industry 4.0 applications} & 
    \textbf{Key Inferences}\\ \hline
    \multirow{4}{*}{\parbox{3cm}{Descriptive Analytics}} & Lade \textit{et al.}~\cite{lade2017manufacturing} & Industrial defects identification & Linear and nonlinear relationships among parts, machineries, and production lines are realised with root cause of defect. \\  \cline{2-4}
    & Vernadat \textit{et al.} \cite{vernadat_information_2018}  & Cyber-physical production & Enables flexible manufacturing processes with modern tools and enhances automation processes. \\  \cline{2-4} 
    & Araz \textit{et al.}\cite{araz2020role} & Smart planning of industrial processes  &  Descriptions on business metrics in terms of customer count, growth impact, financial states are aggregated and analyzed. \\  \cline{2-4} 
	& Ciano \textit{et al.}\cite{ciano2020one} & Lean production & Integration of lean production techniques with Industry 4.0 helps to reduce the cost in the production process.   \\ \cline{2-4} 
 	& Marzouk \textit{et al.}\cite{marzouk2019analyzing} & Building information modeling & Performance of construction projects are analyzed and improves the communication with better
 	visualizations. \\ \cline{2-4} 
 	& Hofmann \textit{et al.}\cite{hofmann2017artificial} & Automotive industry & Enhanced customer focus, with better efficiency in product development process. \\ \hline 
 	
  	\multirow{6}{*}{\parbox{3cm}{Predictive Analytics}} & Dicuonzo et 
  	al.\cite{dicuonzo2019risk} & Risk assessment & AI-based risk extraction and assessment are carried out and provides real-time assistance.\\   \cline{2-4}
	& Zonta \textit{et al.}\cite{zonta2020predictive} & Predictive maintenance & Time-based schemes used on the industrial Big Data for monitoring industrial tasks and carrying out predictive maintenance. \\  \cline{2-4} 
	& Nordal \textit{et al.}\cite{nordal2020modeling} & Reference architecture for predictive maintenance & System engineering methods are applied on the architecture developed for the management of industrial machines using predictive maintenance. \\    \cline{2-4} 
	& Sahal \textit{et al.}\cite{sahal2020big} & Analytics in energy and transportation industries &  Support of open source tools for carrying out analytics ease up the process and provides customized solutions.\\   \cline{2-4} 
	& Tiwari \textit{et al.}\cite{tiwari2020sustainability} & Sustainable industrial operations & Provides mature framework for ensuring sustainability in production process. \\    \cline{2-4}
	& Ngo \textit{et al.}\cite{ngo2020factor} & Construction industries & Capability assessment tool predicts the weakness, strength and capability of materials. \\    \cline{2-4}
	& Dinis \textit{et al.}\cite{dinis2020foresim} & Aircraft maintenance & Bayesian networks are used for forecasting maintenance works based on historical data. \\    \cline{2-4}
	& Schmidt \textit{et al.}\cite{schmidt2017context} & Condition monitoring & Decision-making process are carried out using the data from machining operations. \\    \hline
	
	\multirow{4}{*}{\parbox{3cm}{Prescriptive Analytics}} & Vieira \textit{et al.}\cite{vieira_simulation_2019} & Supply chain management & Provides added-value through transformation of industrial Big Data towards acquiring new skills and strategic resources. \\  \cline{2-4} 
	& Groger \textit{et al.}\cite{groger2018building} & Automation of tasks  & Helps to achieve optimal decisions and supports automation with strict time constraints.  \\  \cline{2-4} 
	& Bousdekis \textit{et al.}\cite{bousdekis2020sensor} & Reliable decision process & The time dependent parameters, helps to make the decision process much easier with realistic outcomes.\\ \cline{2-4}
	& Lepenioti \textit{et al.}\cite{lepenioti2020prescriptive} & Business performance improvement & Optimizes decision making in industry realm using prominent methods.\\ \hline
    \end{tabular}
  \end{center}
\end{table*}
\textbf{}

\subsection{Relationship between AI and Big Data in Industry 4.0}
AI and Big Data are among the key modern technologies that have revolutionized different application domains in the modern world. Both the technologies, though are operationally different from each other, have formed an amazing pair providing a diversified set of applications\cite{zhuang2017challenges}. The main reason behind the connection of both technologies is the data requirements of data-driven AI algorithms. The performance and feasibility of AI algorithms in an application are big-time constrained by the availability of quality data. Thanks to Big Data, it is possible to structure, integrate and identify useful patterns in data, which is then used to train AI/ML models providing outstanding prediction and decision-making capabilities. 
In the industry 4.0 framework, all industrial machines and devices in smart industries are connected, and they interact, communicate and learn from each other. As a result, large volumes of unstructured and heterogeneous data. Big Data techniques/technologies help to analyze, process and extract and identify key patterns in large volumes of data, which is then used to train AI models. To fully exploit the potential of the data generated in smart industries, Big Data and AI should be deployed in an integrated manner.

\section{Challenges of AI and Big Data in Industry 4.0}
\label{sec:challenges}
Despite the outstanding performance of AI and Big Data, several challenges are hindering their way towards a successful deployment in Industry 4.0. In the following subsections, we highlight some key data-related challenges, explainability and interpretation issues, and adversarial and other security attacks on AI in Industry 4.0.  

\subsection{Data Related Issues}
Data lies at the heart of AI systems. They permanently affect the way the AI and Big Data methods respond to different applications. Some of the key data-related challenges to the AI and Big Data methods in Industry 4.0 are described below.

\begin{itemize}
\item \textit{Data Availability}: In modern industries, super-computing resources and cloud platforms along with AI models allow us to make sense of raw data in real-time. However, valid and meaningful insights from AI and Big Data frameworks used in Industry 4.0 applications require a sufficient amount of quality data. The industry knows what the customer demands from the voicing gathered from customers through social media and other public sources. These queries, complaints, and opinions are useful information that could be fed into AI models employed in smart industries along with the data available from the machines and IIoT devices employed. However, gathering various modalities of data and channelizing them for AI systems is very challenging. For instance, the collection of such heterogeneous data from different sources/machines in smart industries requires intensive inter-connectivity with the computational platform\cite{lee2019blockchain}. 

The literature shows that improving the information assets has not been generally the priority of a majority of companies. However, better performances have been observed for data-driven companies compared to others\cite{miragliotta2018data}.  In other words, better industrial productivity could be achieved through quality data and improved AI and Big Data frameworks by enabling accurate decision-making processes. Auditing on data availability ensures concrete policies on data sharing among authorized individuals and allows proper usage of it in industrial applications. Such instances are also carried out in pharmaceutical industries\cite{hopkins2018data}. 
    
\item \textit{Data Accessibility}:  Accessibility of data in a prescribed format for AI models not only helps to make accurate decisions but also helps to forecast the requirement and prospects of smart industries. However, streaming the data from the appropriate source, and easy accessibility of the data to the systems without any hindrance is not straightforward. To solve the issue, several attempts have been made in the literature. For instance, Gaba \textit{et al.}\cite{gaba2020robust} developed a lightweight key exchange mechanism for smooth and secure industrial data access. It was tested to ensure the trust on the IIoT nodes, by preventing the nodes from unauthorized attacks. However, the computation of elliptic cryptographic complexities and its impact on industrial data and accessibility need to be considered for ensuring the legitimization of the IIoT nodes in the industrial network of smart devices.
   
\item \textit{Data Quality}: In learning the right patterns and extracting meaningful insights from the data, the quality of the gathered data plays a crucial role.  The higher the quality of industrial data acquired from different smart machines and other industrial sources the better and accurate are the predictions made by an AI algorithm. Restrictions in access to quality data force businesses and policymakers in smart industries to create substandard AI systems. According to Williams \textit{et al.}\cite{williams2020data}, data quality management is mandatory while integrating Industry 4.0 into lean, powering up smart factories using CPS, IIoT asset management, and enhancing the global supply chain. Transparency in product quality can be supervised using ML approaches to detect the outliers in the data for ensuring its quality\cite{brandenburger2020quality4}.

\item \textit{Data Auditing}: Data auditing is one of the key processes to be carried out in data analytics, where data is analyzed for a particular application. In detail, during the process data collected in an environment for predictive analysis is analyzed whether the available data is suitable for the application or not? Moreover, the potential risks associated with the data are identified\cite{ahmad2020developing}. In smart industries, industrial data is collected from different types of machines, which results in a large volume of unstructured and heterogeneous data. To fully exploit the potential of the data in different applications of industry 4.0, data auditing is essential to analyze and assess the quality of data to be used for training AI/ML models.    

The literature already suggests some interesting efforts in this regard. For instance, Fan \textit{et al.}\cite{fan2020dredas} proposed a decentralized auditing scheme namely Dreads for auditing data collected in an industrial environment. The proposed method ensures three different types of data auditing namely public, dynamic, and batch auditing, and brings several benefits compared to the traditional data auditing schemes.  Wu \textit{et al.}\cite{wu2020convergence} addressed a critical infrastructure in Industry 4.0 while auditing the industrial data. The authors also provide a detailed analysis of the convergence of edge computing and blockchain technology for scalable and secure data management frameworks.  

\item \textit{Data Transmission (Platforms, lack of universal standards, etc.,)}:  Interoperability of data transmission across a diversified range of communication sources remains highly challenging due to the lack of universal standards and common communication platforms. Sharing the data among smart industrial subsystems in meaningful ways through the available infrastructures, such as Ethernet, fiber, or wireless means for the end receiver could be able to interpret the transmitted data and gain meaningful insights from it. 
    
To deal with such issues, Cotet \textit{et al.}\cite{cotet2020innovative} presented a secure, quick, and efficient means of transferring data from the smart industrial sensors to the cloud services. Based on the data collected from waste management systems in industries, it helps to de-clutter the waste from smart factories in a much more effective way. Silvestri \textit{et al.}\cite{silvestri2020maintenance}, on the other hand, provides a detailed overview of data transmission schemes for remote maintenance in Industry 4.0. As remote maintenance becomes a serious concern due to the bloom of smart industrial machines and augmented reality, the proposed schemes help in the self-maintenance of smart factories by minimizing the operators' management tasks.
\end{itemize}

\subsection{Explainability and Intrepretability}
Despite having a slight difference in the concept, explainability and interpretability are used interchangeably in the literature. Interpretability represents the ability of an AI model to link a cause/reason with a decision while explainability on the other hand shows the ability of a model to justify its decision. Both the concepts aim to make AI decisions understandable for humans, which ultimately helps in developing stakeholders' trust in predictions/decisions made by AI in different applications domains\cite{ahmad2020developing}. The interpretation of AI decisions is generally not desirable in low-risk applications. For instance, low risk is associated with an error in movie recommendations. However, interpretations are highly desirable in high-risk applications, such as predictive policing and AI-based recruitment, etc.,  where critical decisions are made based on insights provided by an AI model.  

As detailed in the earlier sections, in Industry 4.0 AI models are used for several tasks, such as predictive maintenance, and predictive quality and yield, etc., where the models need to take critical decisions. In such high-risk applications, interpretability and explainability are highly desirable to reduce the associated risks and increase users' trust in the models. The literature already reports some efforts in this direction. For instance, Rehse \textit{et al.}\cite{rehse2019towards} provides a detailed analysis of the challenges, opportunities, and potential applications of explainable AI in Industry 4.0. Moreover, the authors also illustrate the potential use of explainable AI in process prediction through experimentation in the DFKI-Smart-Lego-Factory prototype. Galanti \textit{et al.}\cite{galanti2020explainable} also proposed an explainable AL solution for process monitoring in the Industry 4.0 setup. The proposed solution is evaluated on real industry benchmark datasets. The traditional black-box AI algorithms could be made explainable either by providing explanations of the provided results or developing explainable and interpretable AI solutions for Industry 4.0 applications. However, there is a trade-off between accuracy and explanations, which is a challenging task to keep a balance between explanations and accuracy of a model\cite{guidotti2018survey}. 
\subsection{Adversarial ML}   
\begin{figure}[!ht]
  \centering    
  \includegraphics[width=0.5\textwidth]{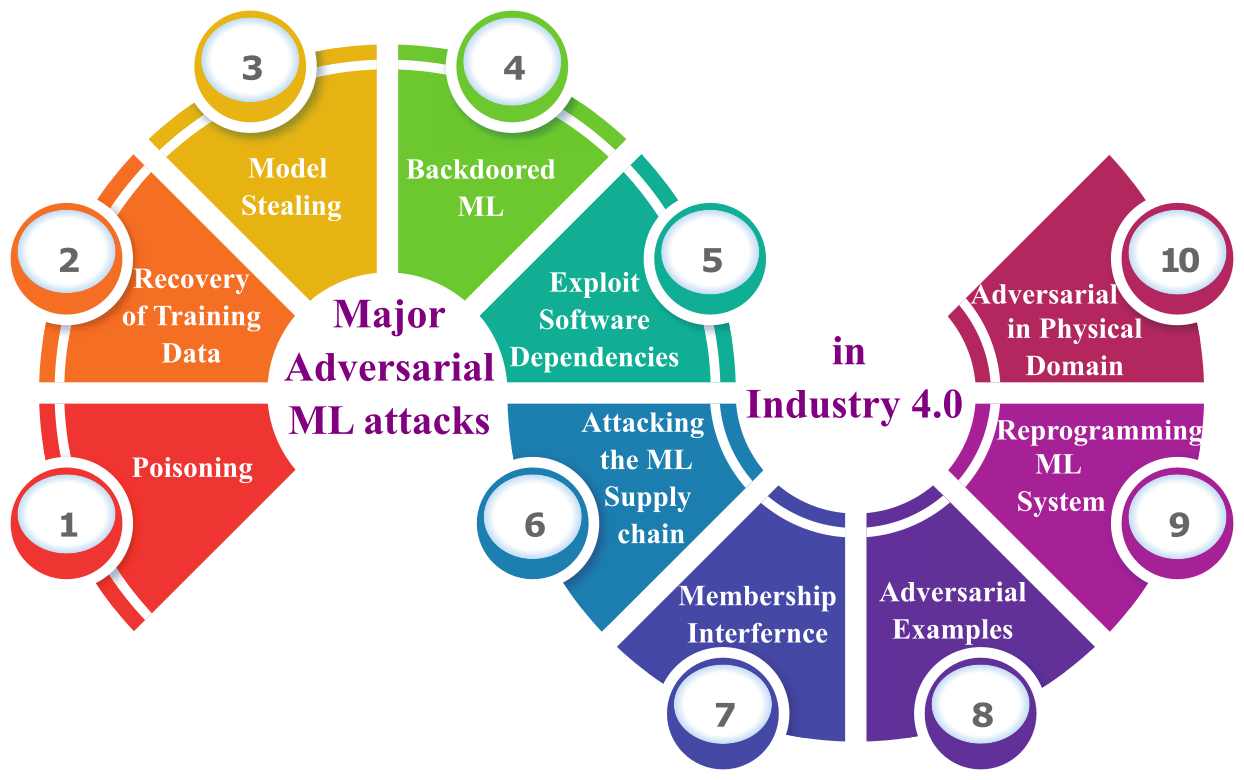}
  \caption{Adversarial ML attacks affecting industrial organizations.}
     \label{fig:adv-ML-attacks1}
\end{figure}

In 2019, Gartner, a leading industry market research firm, published its first report on adversarial ML advising that application leaders must anticipate and prepare to mitigate potential risks of data corruption, model theft, and adversarial samples\cite{kumar2020adversarial}. Figure \ref{fig:adv-ML-attacks1} depicts a high-level overview of common adversarial ML attacks affecting industrial organizations and production control.

A list of adversarial ML attacks is outlined in\cite{kumar2019failure}, from which the top four attacks with the highest impact from an industrial perspective can be outlined as follows.
\begin{itemize}
    \item \textit{Poisoning attacks:} In these attack, adversaries aim to share manipulated labels such that the ML model would consider the re-training process to degrade the performance of the implemented model (assuming that adversaries either have complete control over the training dataset or can contribute to the training dataset\cite{dunn2020robustness}.
    
    \item \textit{Evasion attacks:} Unlike poisoning attacks, evasion attacks may occur after the model training phase, where the adversary may not know about the necessary data manipulation to attack the ML model. In such an attack scenario, adversaries can aim to continuously query the models based on a trial and error fashion, and as result, they can sense how to elaborate the corresponding inputs to pass the ML model. As a consequence, such conduct can lead to resource overhead and impact the system availability and power consumption (especially when devices of limited energy are deployed)\cite{ahmad2020developing}.
    
    \item \textit{Trojan attacks:} In these attacks, adversaries aim to adjust the ML model weights, while maintaining its structure. Although such attacks operate efficiently on normal samples, a trojan target label can further be predicted if a trojan trigger is activated for an input sample\cite{liu2020survey}.
    
    \item \textit{Model extraction/stealing attacks:} In these attacks, adversaries intend to reconstruct or clone the targeted model to compromise the properties and nature of the training dataset\cite{juuti2019prada}. Unlike previous adversarial attacks, these attacks do not require any knowledge about the ML model's architecture or the training dataset. However, adversaries herein are assumed to have access along with its submitted queries' replies\cite{yang2020patchattack, xu2020adversarial}.
    
\end{itemize}

Moreover, a mismatch between Industry 4.0 expectations and existing solutions remains significant when it comes to adversarial ML. Industrial organizations greatly rely on the assumption that solutions available in widely-used platforms (e.g., TensorFlow, PyTorch, Keras, etc.) are inherently robust against adversarial manipulation attacks that have been battle-examined against these attacks. Therefore, organizations typically deploy ML as a service (ML as a Service such as Microsoft Cognitive API), while pushing the security upstream to Service Providers (SP)\cite{kumar2020adversarial}. 

Adversarial ML attacks can lead to severe effects over industrial production control\cite{fawaz2019adversarial}. Although various solutions feeding on texts, images, and videos have been proposed for the smart waste and smart agriculture industries, other smart industrial sectors remain susceptible to a broad range of unintentional ML threats\cite{ahmad2020developing}.

\subsection{Security and Privacy in Industry 4.0}

\begin{figure*}[ht]
\begin{subfigure}{.45\textwidth}
  \centering
  \includegraphics[width=.6\linewidth]{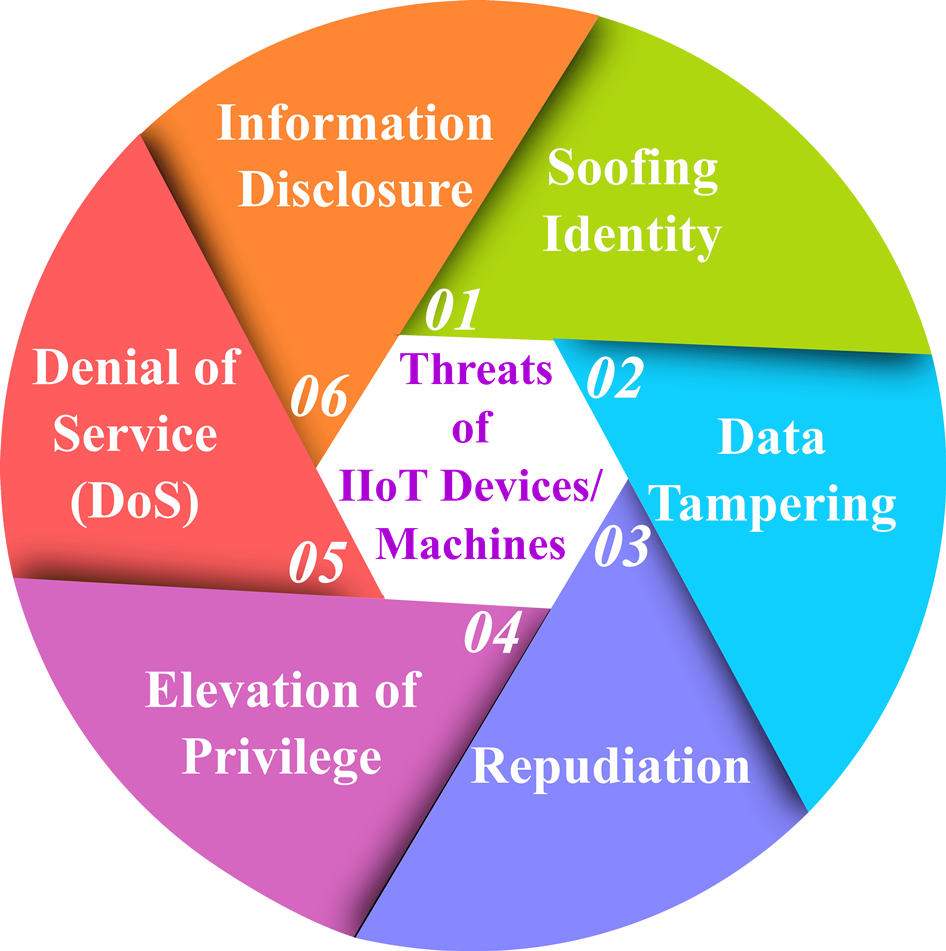}  
  \caption{Common security threats of IIoT devices.}
  \label{fig:IIoTthreats}
\end{subfigure}
\begin{subfigure}{.6\textwidth}
  \centering
  \includegraphics[width=.45\linewidth]{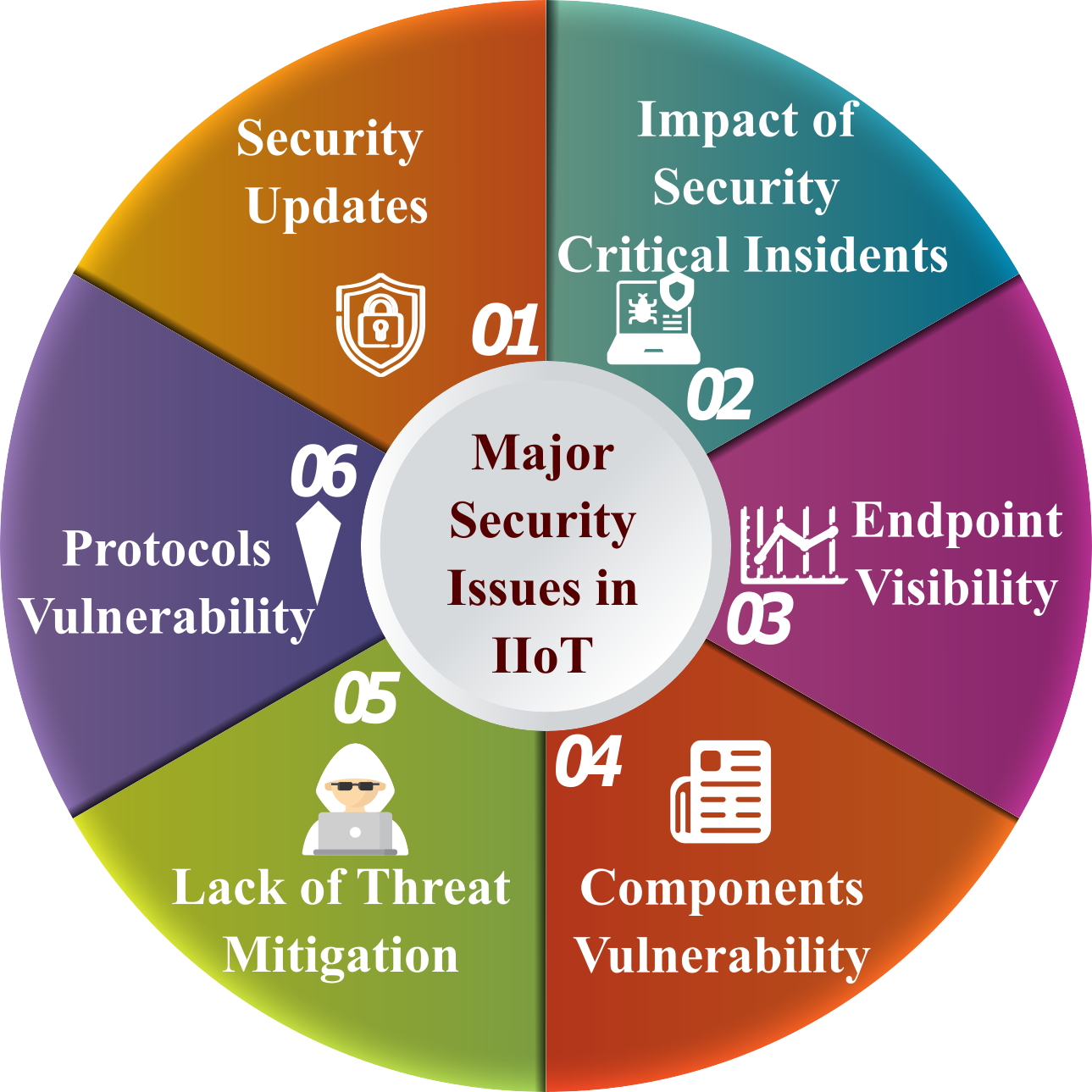}  
  \caption{General security challenges of IIoT environments in Industry 4.0.}
  \label{fig:IIoTsecurity}
\end{subfigure}
\caption{A Taxonomic overview of key security challenges in IIoT.}
\label{fig:IIoT-security}
\end{figure*}

As depicted in Figure \ref{fig:IIoTsecurity}, security in IIoT systems poses several issues including, but not limited to, the lack of cyber risk mitigation, lack of endpoint visibility, security updates, protocols vulnerability, the vulnerability of multiple industrial components when shifting from closed to open networks\cite{aceto_survey_2019}.

\subsubsection{Security Solutions in IIoT Ecosystem}

IIoT ecosystems exploit IoT communication paradigms in various industrial applications, converging toward interoperability and reliability within connected machines \cite{younan_challenges_2020}. The amount of devices (i.e., objects) connected to the Internet has been dramatically rising, and thus IIoT becomes a dynamic network of networks. Various challenges include but are not limited to resiliency, reliability, and security, rendering services in IIoT ecosystems inaccurate or incorrect. These challenges are critical in a broad range of IIoT applications of industry \cite{younan_challenges_2020}. Enabling real-time AI-based solutions for knowledge extraction can reduce cyber risks IIoT systems and control data sharing across the industrial application. 

Furthermore, as IIoT ecosystems keep growing by diverse usage of at-scale measurement devices, IIoT-enabled predictive maintenance (PdM) has become critically necessary in smart manufacturing. Industries have significantly invested in the PdM to improve both the machine parts and the uptime of equipment while optimizing maintenance and security costs\cite{he2017integrated}. However, fault detection remains a prominent challenge in PdM in IIoT-enabled smart manufacturing. Reliable monitoring of such faults requires the involvement of the entire ecosystem from data collection and storage to results delivery (to target applications), rather than ML being solely deployed in the prediction phase\cite{yu_global_2020}. Injection procedures can be integrated with data lake structure in real-time IIoT-based smart manufacturing to enhance fault detection in PdM.

Wireless sensor network (WSN) has also emerged as a substantial component of the contemporary IIoT ecosystems\cite{li_review_2017}. Typically, sensor nodes try to sense phenomena and report findings in their environments to a central base station. These sensor activities enable the core platform for heterogeneous service-centric industrial and business applications\cite{rao_impact_2018}. Yet, security vulnerabilities and threats to sensor nodes in IIoT ecosystems remain considerable. Namely, denial of service (DoS) and distributed denial of service (DDoS) with a significantly large number of networking packets are considered sophisticated attacks against sensor nodes in the IIoT network\cite{baig_averaged_2020}. Smart detection and prevention mechanisms can be elaborated with proper ML techniques\cite{jagannath2019machine}. To strengthen the threat detection accuracy, it is recommended to test the framework over real-world IIoT attack scenarios before deployment\cite{baig_averaged_2020}. 

Furthermore, with the increasing number of connected devices, IIoT ecosystems are expected to have more demands for resilient and reliable communication infrastructures\cite{li_review_2017}. Besides being able to learn how to autonomously retrieve spectrum knowledge on-the-fly, IIoT devices must also utilize this knowledge to securely adjust wireless parameters synchronously to achieve the optimal functionality and operation of the IIoT underlying network using ML mechanisms\cite{jagannath2019machine}. Lastly, Industrial cyber-physical and IIoT systems are the primary enablers of data-driven paradigms and manufacturing intelligence in Industry 4.0. To attain reliable real-time pervasive operational data streams and networks, these industry data-driven facilities are better assessed with ML models, e.g., Predictive Model Markup Language (PMML) encoded ML models in factory operations to assess security and privacy\cite{o2018fog}.

\subsubsection{Privacy in IIoT Cloud Services}

By leveraging ML and DL techniques, industrial artificial intelligence (IAI) has mainly been deployed to address a broad range of industrial limitations and challenges, such as privacy and data sharing, in IIoT-enabled cloud services in Industry 4.0\cite{hao_efficient_2019}. Traditional centralized knowledge training is often inefficient for critical data-driven scenarios, e.g., the case of healthcare\cite{mutlag_enabling_2019}.

Federated learning, an ML mechanism that has recently started emerging and drawn significant attention \cite{yang2019federated}. It trains the prediction model over multiple distributed edge entities carrying unexchangeable local samples of data while allowing sharers to impart a shared model in real-time without the need for revealing their local data samples. Nevertheless, the shared parameters among the local participants can be exploited to compromise applications and cloud services in IIoT platforms (e.g., healthcare data, decision-making-based industrial robots, and navigation services) \cite{qayyum2021collaborative}. A federated learning-based privacy-preserving solution can be designed for IAI to restrain private and sensitive data from leakage and local participants from collisions\cite{hao_efficient_2019}.

Although service-based architecture and cloud computing services have been proven to be a vital integration with cyber-physical and IIoT systems, they are operating based upon information technology perspectives. This raises concerns related to security, privacy, and real-time performance. Therefore, centralized cloud computing implementations can be integrated with decentralized fog computing paradigms to provide privacy-enhanced, real-time, and dynamic ML applications for Industry 4.0\cite{odonovan_comparison_2019}.

The integration of blockchain technology and AI has recently received strong attention for adoption in IIoT applications and services\cite{salah2019blockchain}. However, designing secure and resilient smart contracts for applications remains a grand challenge because of these contracts' complexity\cite{gupta2019tactile}. Smart contracts as self-executable and verifiable can enhance the privacy and security of Industry 4.0 while eliminating the need for a third party implementation\cite{budhiraja2019tactile}. Although such smart contracts yield reforming Industry 4.0, challenges in security and privacy remain significant\cite{salah2019blockchain}. Various privacy vulnerabilities within the software code may be exploited by an adversary to compromise the underlying IIoT networks, whereas developing complex smart contract paradigms cannot guarantee a secure and privacy-enhanced IIoT system\cite{gupta_smart_2020}.

Privacy and security of smart contracts can be enhanced by splitting them into off-chain and on-chain contracts, such that contracts whose operation requires high computation cost and data sensitivity are integrated as off-chain and executed and signed only by the participating entities\cite{li2019scalable}. Furthermore, assessing the influential factors in building an efficient IIoT environment is crucial from the Industry 4.0 perspective. A rule-based decision making approach can be elaborated to better assess the system influential factors\cite{ly_fuzzy_2018}. As security and connectivity characteristics are key tangible influential factors and more significant than intangible factors, it is essential to incorporate proper security schemes with these influential factors while building an enterprise IIoT system\cite{ly_fuzzy_2018}.


\subsubsection{Blockchain security solutions}

Blockchain is a disruptive technology that provides secure, peer-to-peer connectivity and eliminates the need for a central authority. Blockchains provide cheap and faster means of business transactions and eliminate the need for third-party intervention. Blockchain technology enables records to store all transactions in a decentralized ledger. Blockchain safeguards smart industrial machines from cyberattacks and supports effective supply chain management. Also, Blockchain technology is playing a leading role in the healthcare sector, maintaining government records and insurance records with high-end security features. 

\begin{figure}[!ht]
  \centering    
  \includegraphics[width=0.45\textwidth]{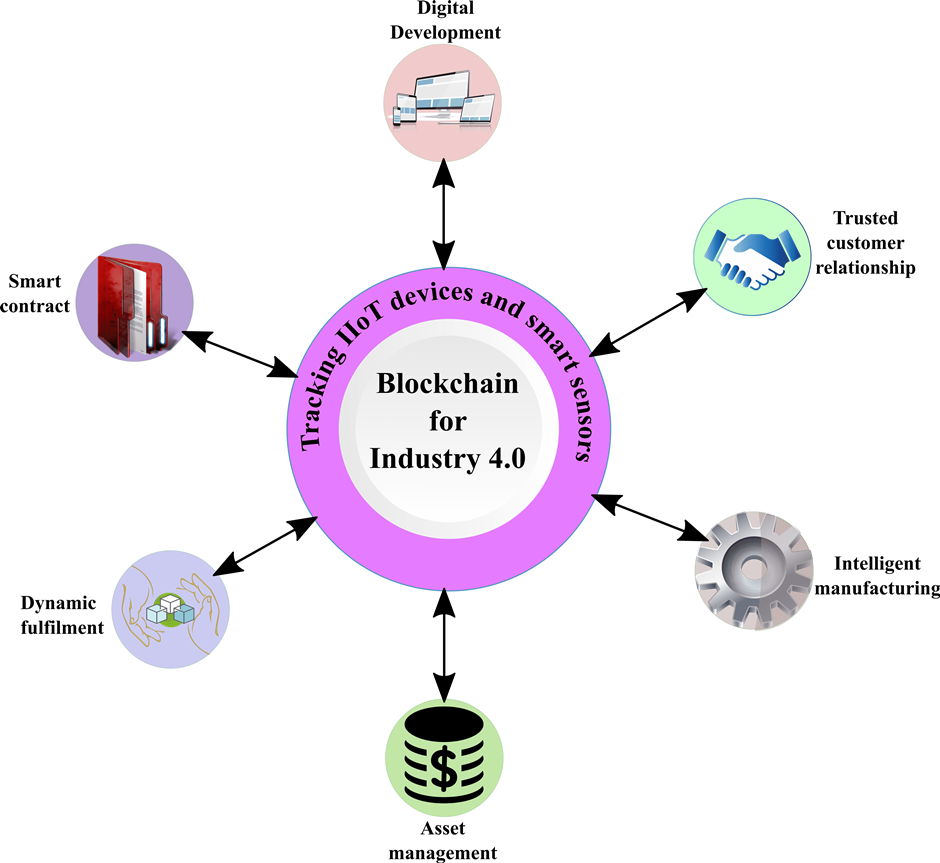}
  \caption{Role of Blockchain with the support of IIoT in Industry 4.0 for Digital supplychain.}
     \label{fig:blockchain}
\end{figure}

Edge intelligence IIoT applications empowered with blockchain support are integrated with cloud services for distributed sharing of industrial resources with reduced cost\cite{zhang2019edge}. The most relevant roles of blockchain along with the support of IIoT devices are summarized in Figure \ref{fig:blockchain}. By taking a closer look at the presented figure, it is clear that there is an indisputable impact of blockchain in supply chain management with the phases of evidence from intelligent manufacturing, asset management, provision of smart contracts and imparting trusted customer relationships. 

\subsubsection{Supply chain security solutions}

Data analytics on IIoT data through the support of ICT enhances the smartness of IIoT devices providing promising enhancements in smartness and intelligence of the systems\cite{seyedghorban_supply_2020}. This framework enables a secured supply chain and providing real-time analytics by leveraging ICT.

Digitization of the food supply chain in industries is carried out based on demand for resources with a secured framework integrated with the IIoT devices\cite{kittipanya-ngam_framework_2020}. Impact of Industry 4.0 with enhancement in the agriculture sector is adapted as agriculture 4.0 by Lezoche \textit{et al.}\cite{lezoche_agri-food_2020}. The authors reviewed and analyzed the secured supply chain methods incorporated by considering several key aspects in the domain of agriculture.

Based on the cascaded review performed by Parente \textit{et al.}\cite{parente_production_2020}, the authors focused their discussions on the scheduling of resources in Industry 4.0 frameworks. Moreover, the authors highlighted the importance of holistic scheduling mechanisms and the flexibility imparted with the aid of appropriate scheduling strategies. 

ML and DL models are proposed by Gupta \textit{et al.}\cite{gupta_machine_2020} to identify unknown attacks on Big Data. The developed threat model analyzes various vital parameters such as reliability, accuracy, latency, and efficiency. This secure data analytics model favors securing the supply chain from threats. 

\subsubsection{Integrated security for maturity models}

In augmentation with human intelligence, collaborative cognition has been adapted and applied to the developed models for appropriate decision making in industrial applications\cite{jiao_towards_2020}.  The maturity model includes cyber-physical-human analysis using novel strategies with an enhanced level of trust and an optimized level of human interactions and automation.

For driving the smart agricultural industry in terms of enhancement in food security, autonomous pollination is developed for farming\cite{chen_intelligent_2019}, by utilizing robotic systems that act as micro air vehicle pollinators. With the support of AI and humans in the loop, the automated systems provide a matured model with enhanced security features. 

CPS-based automation for Industry 4.0 along with IIoT devices are utilized for smart transportation systems that integrate AI, ML, and human-machine interaction support\cite{kumar_novel_2020}. Various bus scenarios are considered to test the modular features of the framework under traffic conditions in a simulation environment. 

Smart-troubleshooting is introduced for collecting the failure status of IIoT devices connected for the implementation of Industry 4.0\cite{caporuscio_smart-troubleshooting_2019}. Further, it is also involved in prognostics and diagnostics of the devices integrated through the smart infrastructure that are manufactured by different vendors. Also, product information, faults encountered and other security threats are also intimated by the smart troubleshooting framework. 

Effective decision making with the aid of data acquired from IIoT devices is carried out using the Maximum Mean De-Entropy technique considering a specific threshold\cite{singh_integrated_2020}. Also, they are used in Interpretive Structural Modelling considering the barriers and challenges faced by the manufacturing industry.

With the application of the Industry 4.0 maturity model\cite{wagire_development_2020}, maturity levels of various manufacturing industries are validated. Moreover, the model was also used to observe the maturity score for self-assessment and real-life implementation in the Industry 4.0 based supply chain operations. 

Maturity levels are assessed from 12 different defense sector-related manufacturing firms that have implemented Industry 4.0 frameworks by Bibby \textit{et al.}\cite{bibby_defining_2018}. The maturity levels are observed by considering eight different attributes of smart manufacturing industries. 

Policies put forward for implementation of Industry 4.0 are addressed by Sung \textit{et al.}\cite{sung_industry_2018}, by considering the stability, security, and reliability issues in the machine to machine communication. The authors summarized the detailed action plans for making innovative changes and providing necessary sophisticated infrastructure for the initiatives towards successful implementation of Industry 4.0 solutions.

\subsubsection{Securing communication between smart industrial devices}

Security threats such as DoS attacks against IIoT devices (i.e., target a node of the industrial network) are still prevalent and can cause catastrophic effects while emergency and crucial operations are carried out. An intelligent DoS detection framework is proposed by Baig \textit{et al.}\cite{baig_averaged_2020}, which was experimented with actual IoT attacking scenarios providing promising accuracy in the classification of DoS attacks. Apart from the adoption of blockchain for transport and logistics, there is also the importance of tracking the mobility of goods through a seamless flow of uninterrupted communication. The role of 5G communications plays a significant part along with blockchain and even supports in establishing secure communication across borders\cite{koh_blockchain_2020}.

\section{Insights, Future Research Directions and Open Issues}
\label{insights}

\subsection{Future aspects of Industry (i.e., beyond 4.0 )}
\label{beyond}
This section focuses on key innovations and implications in modern smart industries that go beyond Industry 4.0 towards Industry 5.0. One of the important key features of adopting Industry 4.0 is the internal cultural transformation in the industry. This requires strong leadership with a commitment to change the management process, invest in the necessary technology enablers and education to successfully adopt smart manufacturing practices. All these tend to be huge investments from the perspective of owners and managers, but the significant operational cost improvements pay them back with huge benefits within a couple of years. With the deployment of Industry 4.0 practices in an Industry, they can compete with anyone in acquiring and retaining potential customers with enhanced operational efficiency. 

Smart manufacturing drives the competitiveness of manufacturing enterprises. Competitiveness is a critical factor, where the manufacturers around the world seek a real-time understanding of the operational status of every machine that is producing any part or component ending up into finished manufactured products.
\subsubsection{Industrial Innovation and Infrastructure}

As trust in the automated decision making by smart industrial machines using AI is getting a serious concern, the demand for trustworthy machines is gaining importance. Explainable artificial intelligence (XAI) has emerged as a transparent mechanism for predicting the requirements in the automated industry with complete exploitation of Industry 4.0\cite{rehse2019towards}. 

For context-aware and ubiquitous computing through the data collected from IIoT devices or sensor nodes, usage of cloud and fog computing helps towards the sustainable transformation of present smart industries\cite{abdel-basset_novel_2019}. 

The impact of IIoT will be significant for critical industrial sectors that address manufacturing, power, and chemicals as the smart industries get their transformation towards Industry 5.0 and beyond. IIoT devices also improve production efficiency, reduce the cost for manufacturing, provide better status monitoring and security-enhanced solutions. Moreover, the blockchain solutions also detect authenticated usage of software and its update management in the IIoT devices\cite{he_bosmos_2020}. 

The miniaturization of modern devices, wearable electronics, high-speed computing, efficient data analytics has opened up avenues for policymakers for transforming the existing industrial standards. The business management, researchers, and technical experts have started visualizing the transformation from technical to socio-technical perspective for efficient and smart manufacturing in future industries\cite{elbanna_search_2020}. 

\begin{figure}[!ht]
  \centering    
  \includegraphics[width=0.45\textwidth]{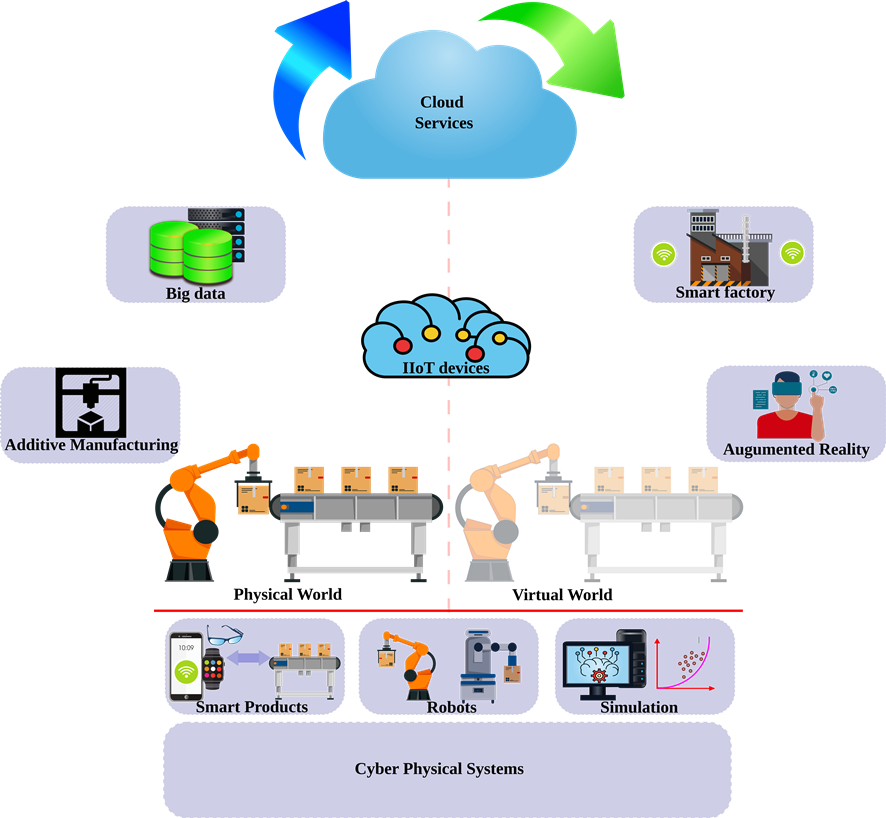}
  \caption{Impact of CPS integrating the physical and virtual world in Industry 4.0 applications.}
     \label{fig:CPSInd40}
\end{figure}

Figure \ref{fig:CPSInd40} illustrates the smart manufacturing process through the integration of the physical and virtual world with the aid of CPS. Robotic systems involved in the production process in Industry 4.0 enhances flexibility, productivity thereby reducing the risks of fatigue and injury for humans. Certain key aspects of the CPS is modern industries are pushed to the simulation part and carried out through virtual systems. IIoT devices and smart machines are integrated with cloud services, Big Data, and additive manufacturing providing enhanced services to the end-users through smart devices and augmented reality. Minimization of interaction forces between humans and robots optimizes the production process and improves the qualitative performance using robust controllers\cite{roveda_model-based_2020}.  

Deployment of high-performance robotic systems in Industry 4.0, enables the smart industries transformation towards Industry 5.0. Involved production process and service roles of the industrial systems are personalized towards customer specification smart production processes. However, those smart industries also address the organizational analysis through socio-technical approaches\cite{bednar_socio-technical_2019}. 

Transparent architecture with user-friendly and readable coding processes makes novice programmers program and control industrial devices and control them. Such illustrations of the conceptual model drive smart industrial objects to adapt to the modern industrial environment and paves way for effective means of industrial transformation\cite{reenskaug_personal_2019}. 

\subsubsection{Sustainability in Industrial Revolution}

The research community, academicians, and business experts have defined many road maps for sustainable implementation of Industry 4.0 considering the societal and economic progress through a systematic approach\cite{oztemel_literature_2020}. 

Robotics for domestic services and industrial automation tasks along with AI and image vision have enough expertise towards the adaptation of Industry 5.0. Moreover, the enhancement in terms of production quality, improved efficiency, and reduced cost benefits can be addressed with auto-adaptation for providing quality services\cite{massaro_re-engineering_2020}. 

Over digital networks, automation built using robotic systems for connecting the urban sector through the industrial revolution is addressed through the socio-spatial perspective through innovation and speculative modernization and transformation of smart industries\cite{macrorie_robotics_2019}.  

Automation of smart industrial objects through IIoT along with the enabling technologies and challenges faced are highlighted to envision the revolution of industries with substantial development towards Industry 5.0\cite{khan2020industrial}.  

Sustainable Development Goals at the international level are well established by addressing the economy and ecology for the industrial revolution with sustainable growth in production and automation through innovative technologies. The economic y transformation process is hindered through the technology towards the successful deployment of the smart factories for sustainable development\cite{schutte_what_2018}. 
\subsubsection{Re-configurable factories}

Manufacturing processes can be made adaptive and reconfigurable based on consumer demands and expectations. The product cycle of goods is tailored to the requirements of customers with improved serviceability and scalability with affordable cost for the consumers\cite{brad_design_2018}. 

\subsubsection{5G Private Networks for Industry 4.0}
\textcolor{black}{Nowadays, the usage of 5G technology, including its private networks, is considerably increasing \cite{cheng2018industrial}. As a result, manufacturers will have to ensure their network technologies are functioning at maximum capacity while leveraging the 5G private networks into their AI-enabled production systems and applications. Notably, various integrated solutions can adopt the capabilities of 5G technology to bridge the noticeable gap between the manufacturing networks and the private Long-Term Evolution (LTE) \cite{varga20205g}. The private 5G networks here can supplant the deployment of WiFi in industrial production settings (and applications) through the Citizens Broadband Radio Service (CBRS). Such a novel and adaptive wireless connectivity feature can fit the various AI and big data-based industrial applications, which require real-time and uninterrupted co-operation/interoperability to perform decision-making, e.g., IIoT devices that require ultra-reliable networks for an efficient time-sensitive data collection and transfer.}

\textcolor{black}{Furthermore, private 5G networks can provide the desired latency for future AI-enabled smart manufacturing applications in Industry 4.0, which is a critical factor for automation processes over big data management and analytics \cite{aijaz2020private}. Lastly, private mobile networks can be regarded as remarkable and new networking architecture extensions to bridge the current gap between AI, smart manufacturing systems, and edge computing. Information applications of Industry 4.0 and big data analytics still reside within the onsite data center or the cloud; however, the adoption of private 5G networks can potentially extend the functionality of AI applications to the manufacturer's floor for better automation of industrial settings and systems, providing reliable and predictable inter-communication between guided vehicles, system administrators, and AI-based decision-making applications.}

\subsection{Open Issues}

This sub-section provides sample instances of a few potential research directions for the researchers working on AI and Big Data and their implications towards Industry 4.0. Some key potential future research directions in AI and Big Data-based solutions for Industry 4.0 are : 

\begin{itemize}
    
    \item \textbf{Energy constraints:} Energy and hardware constraints are among the significant and crucial challenges hindering the deployment of ML in Industry 4.0. The research contribution is therefore needed to enhance and optimize energy consumption and conservation in IIoT devices\cite{younan_challenges_2020}. Real-time ML, a research trend that has recently emerged with the hope of addressing these challenges through online or incremental learning, could be a potential direction of future research. Incremental learning is perceived to replace the traditional data training module with incremental-based training, improving both energy and hardware (e.g., memory) consumption.
    
    \item \textbf{Scalability issues:} Scalability is another significant challenge that may also be considered for future research. The number of IoT devices is expected to reach several trillion in the next few years, and thus, performance and information monitoring challenges will increase remarkably\cite{younan_challenges_2020}. Besides, as the number of IIoT devices vastly increases, privacy and security must be reconsidered. Namely, ML and/or real-time ML solutions can be developed to thwart run-time attacks and identify various security threats at both the user and network levels. Besides, management operations such as device discovery, enumeration, and firmware updates must be taken into account while designing and placing any security solution over IIoT environments. Lastly, the legitimacy of geolocation data shared by IIoT devices is an additional research challenge. A few studies in the past aimed to resolve this challenge using covert channels (e.g.,\cite{islam2017determining}), but the challenge of trust of affected third parties and privacy of direct users remains an open research problem as a trade-off between security and privacy-preserving geolocation-enabled IIoT devices.

    \item \textbf{Cloud Computing:} Cloud computing and performance is an additional vital component in Industry 4.0-enabled cloud multi-faced paradigms. Although this provides a baseline mechanism for handling the on-demand metered services, cloud computing, as well as communication protocols performance and efficiency, pose limitations for Industry 4.0 objectiveness. Such limitations include, but are not limited to, the criticality level of the underlying communication technology linking the cloud with clients under dynamic and unpredictable end-to-end delay conditions. As a result, an open research problem herein is designing AI-based sustainable solutions to assist in predicting cloud-user communication latency while preserving energy consumption and recommended QoS. Besides, the design of efficient ML-based monitoring mechanisms for latency-sensitive applications, such as emergency response systems and virtual reality, requires consideration of the large and pervasive amount of information generated by the connected devices\cite{aceto2020industry, greco2020trends}.
    
    \item \textbf{Edge Computing:} Edge computing is the other important aspect and enabling technology that is expected to fully explore the potential of AI in Industry 4.0 \cite{sitton-candanedo_review_2019,trinks2018edge}. Edge computing enables to process data locally (i.e. near to the collecting devices) resulting in a significant reduction in bandwidth and latency. However, there are several challenges associated with edge computing hindering the successful deployment of AI algorithms in Industry 4.0. For instance, the computational and storage resources scarcity of edge devices is the biggest challenge to successfully deploy DL techniques at the edge. One of the potential solutions is the design of computation and memory-efficient DL architectures. Network pruning and distillation are the other potential solutions to cope with the computational and memory issues in deploying DL algorithms for Industry 4.0 applications at the edge.

    \item \textbf{Robotics:} Recently, several tentative robotics-based application proposals to new industrial scenarios and services lead to newly evolving research problems, e.g., DARPA robot challenges\cite{rodic2018smart}. Precisely, emerging robotics-based scenarios in Industry 4.0 create novel and critical challenges related to efficiency, resiliency, and security. These challenges are due to teleoperation and/or human-based cooperation, traffic monitoring, hardware/software maintenance (e.g., firmware update), and real-time framework performance. By leveraging advanced APIs and real-time performance specifications, AI techniques can be of great interest in employment for addressing these emerging robotics-based framework challenges in Industry 4.0. Although human-robot interaction is still regarded as one of the major innovative challenges in Industry 4.0 vision, the human role is irreplaceable sometimes\cite{maly2016augmented}. Therefore, AI-based cyber-physical systems user interfaces must be quietly enhanced or re-designed with specific requirements focused on Industry 4.0 applications\cite{weiss2016first}. Besides, the coordination between robots and humans invokes emerging and innovative research trends due to the high expectation of automation in Industry 4.0.
    
    \item \textbf{Machine-to-machine communication:} Additionally, intelligent machines, along with conveyors in Industry 4.0, are intended to inter-communicate with each other to maximize production lines. Decision making, action triggering, and independent control are further enabled in Industry 4.0 production systems (i.e., smart manufacturer). Underlying networks in Industry 4.0 must, therefore, capture a tremendous amount of machine data to the corresponding cloud. Hence, designing ML-based system-wide feedback and cooperation among these smart objects/machines can be emphasized by future research directions to enhance the performance of the smart production systems.

    \item \textbf{Cyber risk management:} The Industry 4.0 goals to be fully attained, it is substantial to monitor, analyze, and interact with the virtual and real production systems concerning a high level of precision. Hence, computer vision can be aligned with AI approaches to collect, analyze, and synthesize generated data to attain both security and efficiency in Industry 4.0 infrastructures and solutions. Specifically, key goals in aligning these two disciplines are (1) achieve efficient cyber risk, and threat management specific to Industry 4.0 scenarios and applications, (2) develop dedicated management and control frameworks to characterize critical assets such as passive monitors, and (3) elaborate secure-by-design security schemes for endpoint entities (e.g., IIoT devices).

    \item \textbf{IIoT:} As IIoT plays a fundamental and crucial role in Industry 4.0, their associated risks and issues must be further analyzed, whereas the complexity of the underlying IIoT network is better to split up into three levels before any AI-based implementation; (1) physical infrastructure, (2) communication protocols, and (3) logical topology. Based upon the presented state-of-the-art literature review, the critical research challenges in IIoT include but are not limited to, security and privacy, lack of standardization, and reliability of existing Industrial technologies that support the IoT field, e.g., communication-enabled technologies and intelligent objects design.
    
    \item \textbf{Industrial data management:} The massive IIoT-generated amount of data needs to be seamlessly monitored and collected through ubiquitous sensors to simplify and expedite scenario-based decision making. Up to date, existing data trading and sharing frameworks are still inefficient, where the collected data can solely be processed or analyzed by the data owners (organizational level). This limitation needs to be considered in future research directions as it significantly delimits the IIoT data value, and as a result, restricts scenario-based decision-making platforms in Industry 4.0.

    \item \textbf{Interpretation vs. Performance}: Explainability and interpretation of AI algorithms' outcomes are also one of the potential future research directions in Industry 4.0. We believe, especially considering the risks involved with AI predictions in the industry, the focus of future AI-based solutions in Industry 4.0 will be on explainable AI. Some efforts are already made in this direction, and it is expected to be well explored in near future.

    \item \textbf{Networking and Fog Computing for AI}: A broad range of solutions and framework implementations have been presented in the past to be used in various industrial processes and domains. However, fog and cloud computing remain with remarkable open challenges where intensive research studies are still needed. Specifically, the previously presented solutions rely on the deployment or availability of the factory's physical nodes, while the corresponding platforms and libraries supporting fog and cloud computing are still limited. Moreover, devices in the fog level are usually of modest capabilities, which renders deployment of some ML implementations (with a high computation requirement) inefficient. Therefore, it is required to consider multiple fog-enabled devices to improve the execution efficiency while gaining accurate and efficient analytics on time. Besides, future research directions may further consider fog-enabled parallel ML implementations for industrial Big Data analytics.
    
    \item \textbf{Adversarial ML} As discussed earlier, cloud computing is prone to several security issues where AI/ML models can be attacked in several ways as detailed earlier. To cope with the adversarial attacks on AI/ML, different defensive techniques can be utilized\cite{qayyum2021collaborative}. The development of adversarially robust AI/ML models for Industrial applications is still an open research issue.
    
    \item \textbf{Human-Machine Interaction (HMI)}: Recent research studies have brought up the potential of incorporating and leveraging humans in industrial applications and contexts. However, they also capture the significant complexity within the decision-making processes. Hence, further validation (e.g., using the inferential analysis to identify key differences between developed solutions) of the generic scalability, feasibility, and applicability of the implemented ML frameworks to achieve fully scalable Industry 4.0 human-involved solutions, applications, real-assembly lines, and production setups over the corresponding resources.
\end{itemize}

Table \ref{tab:Challenges} summarizes some key challenges to a successful deployment of AI and Big Data solutions in Industry 4.0 along with potential solutions. 

\begin{table*}
  \begin{center}
    \caption{AI-based Industry 4.0 application challenges and potential solutions.}
    \label{tab:Challenges}
    \begin{tabular}{|p{2cm}|p{2.5cm}|p{6cm}|p{5.5cm}|}
    \hline
    \textbf{Challenges}& 
    \textbf{Issues} & 
    \textbf{Potential impact on Industry 4.0 applications} & 
    \textbf{Future research directions}\\ \hline
    
    \multirow{5}{*}{\parbox{2cm}{Data-related \\issues}} & Data Availability  & The collection of heterogeneous data from different sources/machines in smart industries requires intensive inter-connectivity with the computational platform. & Industries could migrate workloads and leverage using cloud services and gain a complete advantage of them as Platform-as-a-Service (PaaS) resources.\\ \cline{2-4} 
									 & Data Accessibility & Channelized streaming of data with appropriate security measures in an industrial environment is essential for training effective AI models. Failure in accessing quality data in a prescribed format for AI models may harm the performance of the models. & By progressing and enhancing beyond location or infrastructure-based data accessibility to the user- or activity-based data accessibility using AI solutions.\\ \cline{2-4}
									 & Data Quality	 & Failure in access to quality data may force the businesses and policymakers to create substandard AI systems for different tasks in smart industries. & \textcolor{black}{Managing data quality while integrating it with Industry 4.0 based on asset management of the global supply chain.}     \\ \cline{2-4} 
									 & Data Auditing  & Failure in assessing and mitigation of risks associated with data and unauthorized access to them may result in the quality of prediction based on the data. & \textcolor{black}{Design of intrusion detection mechanisms to identify and thwart threats of malignant data and unauthorized access to enhance the prediction quality.}      \\ \cline{2-4}
									 & Data Transmission  & \textcolor{black}{Failure of providing high-speed communication and data transmission may impact truncation of vital information.} & \textcolor{black}{Establishing efficient and resilient data transmission standards for reliable end-to-end QoS communication systems.}   \\ \hline
    
    \multirow{4}{*}{\parbox{2cm}{Adversarial \\Machine \\Learning}} & Poisoning attacks  &  Manipulation of labels in the training data by an attacker allow them to control the prediction capabilities of the model and predict what they want them to do. & \textcolor{black}{Apply data sanitization, remove outliers from the training datasets, and evaluate the legitimacy of data through spectral clustering with similarity metric.}  \\ \cline{2-4} 
									 & Evasion attacks & The manipulation of training data by an attacker may lead to significant loss by making wrong predictions in some key industrial applications. & \textcolor{black}{Deploy dimensionality reduction, adversarial training of ensemble, train dataset through Ensemble Adversarial Training, and augment the training dataset using adversarial examples with different models.} \\ \cline{2-4} 
									 & Trojan attacks  &  Similar to other attacks, Trojan attacks may lead to a significant loss in making wrong predictions. To avoid such attacks, weights of the ML models are needed to be activated through proper activation. & \textcolor{black}{Implement input-based anomaly detection, re-train all models if the NN is accessible by the adversary, and pre-process the inputs before the NN deployment to prevent inputs from launching the Trojan.}  \\ \cline{2-4} 
									 & Model stealing (extraction) attacks & Such attacks target models' confidentiality where the motives could be stealing and reusing the model for their task or the stolen model could be used for some other adversarial attacks. & \textcolor{black}{Deploy parameters optimization and synthesis queries, manipulate the output probability manipulation, delete the probability of some classes and outputs, and use watermarking techniques.}  \\ \hline
    
    \multirow{4}{*}{\parbox{2cm}{Security and \\Privacy}} & IIoT Cloud Services  & IIoT cloud services may raise several concerns related to security, privacy, and real-time performance. Thus, distributed resource sharing of sensitive industrial data should be verified.  & \textcolor{black}{Design of efficient AI-based monitoring mechanisms for latency-sensitive applications, while considering the pervasive amount of information generated by the IIoT devices.}    \\ \cline{2-4} 
									 & Blockchain & \textcolor{black}{The design of secure and resilient smart contracts for applications is a considerable challenge because of their complexity.}  & \textcolor{black}{Trust and intelligence in manufacturing tasks needs to be imparted.}     \\ \cline{2-4} 
									 & Supply chain  & \textcolor{black}{The integration of smart manufacturing chains is required in a collaborative real-time response manner to meet the changing demands of supply chains.} & \textcolor{black}{Exploitation of ICT tools and digitization should be leveraged.}            \\ \cline{2-4} 
									 & Integrated security	  &  CPS solutions and robust IIoT services with cloud services need security enrichment during their integration. &  \textcolor{black}{Implement robust hardware security mechanisms for cyber-attacks defense, especially in infrastructure-less networks and devices.}   \\ \hline
    
    \multirow{4}{*}{\parbox{2cm}{5G \\Communication}} & Wireless services  & Establishment of connected services among the machines in industries. &  \textcolor{black}{Slicing the physical 5G network into isolated logical networks and integrate softwarization.}    \\ \cline{2-4} 
									 & Securing communication &  \textcolor{black}{Enhanced security features are key enablers for faithful industrial automation and yet need attention.} &  \textcolor{black}{Tracking of goods mobility via a seamless flow of uninterrupted communication. Deploy 5G and decentralized ledgers to establish secure communication across borders.}    \\ \cline{2-4} 
									 & Decision support  & Accuracy in decision making driven through 5G services need to maintain the required QoS. & \textcolor{black}{Implementation of adaptive schemes to support explainability and interpretation of AI-based predictions while preserving QoS of 5G services.}                  \\ \cline{2-4} 
									 & Disruption  &  Popularity of CPS and communication among robotic systems and automation tools need attention. & \textcolor{black}{Deploy computer-aided tools and robots along with CPS to deliver flexible manufacturing processes and automation with support of 5G communication service.} \\ \hline
    \multirow{2}{*}{\parbox{2cm}{IIoT Data \\Handling}} & Data Analytics tools  &  Concrete data analytics tools may impact the machine behavior's prediction.  & \textcolor{black}{Manage large datasets using efficient handling and concrete analytics tools that optimize industrial data transfer across nodes and accurate prediction of machine behaviour.}  \\ \cline{2-4} 
									 & Trustworthiness &  Trust in cognitive prediction is in demand for faithful human-machine interactions. & \textcolor{black}{Design of efficient data (collected by distributed sensors) auditing mechanisms to identify heterogeneous and unstructured data.} \\ \hline
    \end{tabular}
  \end{center}
\end{table*}


\subsection{Insights and Lessons Learned}

We obtained the following insights and lessons from this study.

\begin{itemize}
    \item Digitization is a key to success for industries in the future. The speed at which the industries adopt digitization plays a crucial role. Organizations must become more agile to technology disruption that does not depend on technological innovations only but the adaptability of employees to new working environments and processes. To keep up with the constantly changing environment, employees need to be continuously updated with the development of technology.
    
    \item Recent advancements in blockchain security solutions have enabled the surmounting of security limitations inherent in machines available in modern manufacturing sectors. 
    
    \item Operational excellence is a precondition for smart industry applications in the creation of a hundred percent stable, reliable, and predictable processes. Data analytics and application of the six sigma contribute to reducing variations, increasing process capability, and predictive maintenance and contribute to the prevention of and scheduled downtime and zero defects.
    
    \item In the modern industries several sensors from different vendors are used to sense and collect data about different operations, which results in heterogeneous and unstructured data. Several challenges are associated with the resultant data, which need particular attention during data auditing to identify useful data and the risks associated with it. 
    
    \item Adversarial attacks on AI models are very common in the modern world, and they have serious consequences on different operations of Industry 4.0. To guard against adversarial attacks, all the stakeholders should show the responsibility of risk prevention and mitigation\cite{ahmad2020developing}.
    
    \item There are several risks involved with the AI-based solutions in Industry 4.0 where AI models need to make critical decisions. Explainability and interpretation of AI predictions can provide an opportunity to better understand the risks involved with AI-based solutions in industries, which will ultimately help in developing stakeholders' trust in AI-based solutions in Industry 4.0. However, there is a trade-off between performance and explanations. Transparent models are good at explanations while the black box models are better in predictions.

\end{itemize}
\section{Conclusions}
\label{conclusion} 
The literature depicts that future industries will be fueled by AI through the support of smart machines, Big Data, IIoT, robots, high-speed communication architectures, blockchain, and the broader transition of the economy. However, several challenges hinder its way in a successful deployment in Industry 4.0 applications. This paper provides a detailed analysis of existing efforts made for a successful deployment of AI in different applications of Industry 4.0.  In detail, we have reviewed and analyzed the role of AI and Big Data in Industry 4.0 with a particular focus on the key applications, enabling technologies, techniques, and the associated challenges including security, adversarial attacks, communication, interpretability, and data-related issues. We also provided a detailed analysis of the benchmark dataset available for the training and evaluation of AI-based solutions in Industry 4.0. The paper also identifies open research issues in the domain. We believe such a rigorous analysis of the domain will provide a baseline for future research.

\section*{Acknowledgment}
The authors gratefully acknowledge the Management, and Faculty of Mepco Schlenk Engineering College, Sivakasi, India for their support and extending necessary facilities to carry out this work.
\ifCLASSOPTIONcaptionsoff
  \newpage
\fi
\bibliographystyle{IEEEtran}
\bibliography{Indm.bib}
\begin{IEEEbiography}
[{\includegraphics[width=1in,height=1.25in,clip,keepaspectratio]{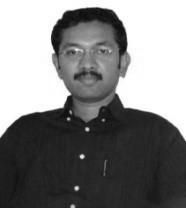}}]
{Senthil Kumar Jagatheesaperumal}
received his B.E. degree in Electronics and Communication Engineering from Madurai Kamaraj University, Tamilnadu, India in 2003. He received his Post Graduation degree in Communication Systems from Anna University, Chennai, in 2005. He has pursued Ph.D. in Embedded Control Systems and Robotics from Anna University, Chennai in 2017. He has 14 years of teaching experience and currently working as an Associate Professor in the Department of Electronics and Communication Engineering, Mepco Schlenk Engineering College, Sivakasi, Tamilnadu. He received two funded research projects from National Instruments, USA each worth USD 50,000 during the years 2015 and 2016. He also received another funded research project from IITM-RUTAG during 2017 worth Rs.3.97 Lakhs. His area of research includes Robotics, Internet of Things, Embedded Systems and Wireless Communication. During his 14 years of teaching he has published 20 papers in International Journals and more than 25 papers in conferences. He is a Life Member of IETE and ISTE.
\end{IEEEbiography}
\begin{IEEEbiography}
[{\includegraphics[width=1in,height=1.25in,clip,keepaspectratio]{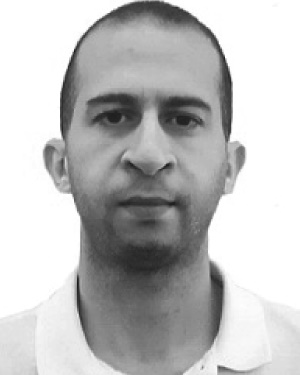}}]
{Mohamed Rahouti} received the M.S. degree and Ph.D. degree from the University of South Florida in Mathematics Department and Electrical Engineering Department, Tampa, FL, USA, in 2016 and 2020, respectively. He is currently an Assistant Professor, Department of Computer and Information Sciences, Fordham University, Bronx, NY, USA. He holds numerous academic achievements. His current research focuses on computer networking, software-defined networking (SDN), and network security with applications to smart cities. 
\end{IEEEbiography}
\begin{IEEEbiography}
[{\includegraphics[width=1in,height=1.25in,clip,keepaspectratio]{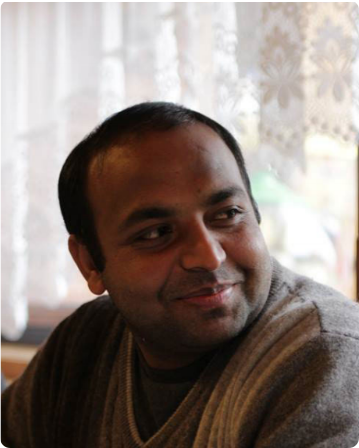}}]
{Kashif Ahmad} received the bachelor’s and master’s degrees from the University of Engineering and Technology, Peshawar, Pakistan, in 2010 and 2013, respectively, and the Ph.D.
degree from the University of Trento, Italy, in 2017. He worked with the Multimedia Laboratory in DISI, University of Trento. He is currently working as a Research Fellow with the Division
of Computer Science and Engineering, Hamad Bin Khalifa University, Doha, Qatar. He has also worked as a Postdoctoral Researcher at ADAPT, Trinity College, Dublin, Ireland. He has authored and coauthored more than 40 journal and conference publications. His research interests include multimedia analysis, computer vision, ML, and signal processing applications in smart cities. He is a program committee member of multiple international conferences, including CBMI, ICIP, and MMSys.
\end{IEEEbiography}
\begin{IEEEbiography}
[{\includegraphics[width=1in,height=1.25in,clip,keepaspectratio]{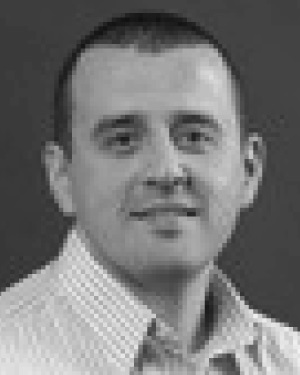}}]
{Al-Fuqaha} [S'00-M'04-SM'09] received Ph.D. degree in Computer Engineering and Networking from the University of Missouri-Kansas City, Kansas City, MO, USA, in 2004. He is currently a professor at Hamad Bin Khalifa University (HBKU). His research interests include the use of machine learning in general and deep learning in particular in support of the data-driven and self-driven management of large-scale deployments of IoT and smart city infrastructure and services, Wireless Vehicular Networks (VANETs), cooperation and spectrum access etiquette in cognitive radio networks, and management and planning of software defined networks (SDN). He is a senior member of the IEEE and an ABET Program Evaluator (PEV). He serves on editorial boards of multiple journals including IEEE Communications Letter and IEEE Network Magazine. He also served as chair, co-chair, and technical program committee member of multiple international conferences including IEEE VTC, IEEE Globecom, IEEE ICC, and IWCMC.
\end{IEEEbiography}
\begin{IEEEbiography}
[{\includegraphics[width=1in,height=1.25in,clip,keepaspectratio]{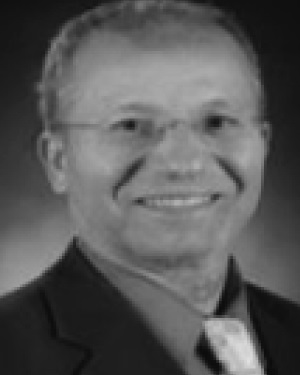}}]
{Mohsen Guizani} (S’85–M’89–SM’99–F’09) received the B.S. (with distinction) and M.S. degrees in electrical engineering, the M.S. and Ph.D. degrees in computer engineering from Syracuse University, Syracuse, NY, USA, in 1984, 1986, 1987, and 1990, respectively. He is currently a Professor at the Computer Science and Engineering Department in Qatar University, Qatar. Previously, he served in different academic and administrative positions at the University of Idaho, Western Michigan University, University of West Florida, University of Missouri-Kansas City, University of Colorado-Boulder, and Syracuse University. His research interests include wireless communications and mobile computing, computer networks, mobile cloud computing, security, and smart grid. He is currently the Editor-in-Chief of the IEEE Network Magazine, serves on the editorial boards of several international technical journals and the Founder and Editor-in-Chief of Wireless Communications and Mobile Computing journal (Wiley). He is the author of nine books and more than 750 publications in refereed journals and conferences. He guest edited a number of special issues in IEEE journals and magazines. He also served as a member, Chair, and General Chair of a number of international conferences. Throughout his career, he received three teaching awards and four research awards. He is the recipient of the 2017 IEEE Communications Society Wireless Technical Committee (WTC) Recognition Award, the 2018 AdHoc Technical Committee Recognition Award for his contribution to outstanding research in wireless communications and Ad-Hoc Sensor networks and the 2019 IEEE Communications and Information Security Technical Recognition (CISTC) Award for outstanding contributions to the technological advancement of security. He was the Chair of the IEEE Communications Society Wireless Technical Committee and the Chair of the TAOS Technical Committee. He served as the IEEE Computer Society Distinguished Speaker and is currently the IEEE ComSoc Distinguished Lecturer. He is a Fellow of IEEE and a Senior Member of ACM.

\end{IEEEbiography}

\end{document}